\RequirePackage{fix-cm}
\documentclass[smallextended]{svjour3} 
\smartqed
\usepackage{array}

\usepackage{mathptmx}
\usepackage[sort&compress,numbers]{natbib}
\usepackage{caption}
\usepackage{mathtools}
\usepackage[hidelinks]{hyperref}
\usepackage[utf8]{inputenc}
\usepackage{graphicx}
\usepackage{amsmath}
\usepackage{booktabs}
\usepackage{float}
\usepackage{amsfonts}
\usepackage{subcaption}
\usepackage{multirow}
\usepackage[bottom]{footmisc}
\journalname{Multimedia Tools and Applications}

\title{LGVTON: A Landmark Guided Approach for Model to Person Virtual Try-On}

\author{Debapriya~Roy,
        Sanchayan~Santra,
        Bhabatosh~Chanda}
\institute{D. Roy \at
              Indian Statistical Institute \\
              \email{debapriyakundu1@gmail.com}           
           \and
           S. Santra \at
              Institute for Datability Science, Osaka University\\
              \email{sanchayan.santra@gmail.com}           
            \and
            B. Chanda \at
            Indian Statistical Institute \\
              \email{chanda@isical.ac.in}\\
}

\date{Received: date / Accepted: date}
\titlerunning{}
\authorrunning{Roy et al.}
\DeclareUnicodeCharacter{2212}{-}
\begin{document}
\sloppy
\maketitle
\begin{abstract}
In this paper, we propose a Landmark Guided Virtual Try-On (LGVTON) method for clothes, which aims to solve the problem of clothing trials on e-commerce websites. Given the images of two people: a person and a model, it generates a rendition of the person wearing the clothes of the model. This is useful considering the fact that on most e-commerce websites images of only clothes are not usually available. We follow a three-stage approach to achieve our objective. In the first stage, LGVTON warps the clothes of the model using a Thin-Plate Spline (TPS) based transformation to fit the person. Unlike previous TPS-based methods, we use the landmarks (of human and clothes) to compute the TPS transformation. This enables the warping to work independently of the complex patterns, such as stripes, florals, and textures, present on the clothes. However, this computed warp may not always be very precise. We, therefore, further refine it in the subsequent stages with the help of a mask generator (Stage 2) and an image synthesizer (Stage 3) modules. The mask generator improves the fit of the warped clothes, and the image synthesizer ensures a realistic output. To tackle the problem of lack of paired training data, we resort to a self-supervised training strategy. Here paired data refers to the image pair of model and person wearing the same cloth. We compare LGVTON with four existing methods on two popular fashion datasets namely MPV and DeepFashion using two performance measures, FID (Fréchet Inception Distance) and SSIM (Structural Similarity Index). The proposed method in most cases outperforms the state-of-the-art methods. 

\end{abstract}

\section{Introduction}
\label{sec:intro}

Buying clothing items online always includes the uncertainty of the fitting and appearance of the item on the customer. Since it is not possible to physically try these items before buying, we are left with the option of trying the clothes virtually. On the other hand, many times we want to find out how we may look in particular clothing that someone else is wearing. Virtual Try-On (VTON) methods aim to address these issues.

\begin{figure}[tb]
\centering
\includegraphics[width=0.7\textwidth]{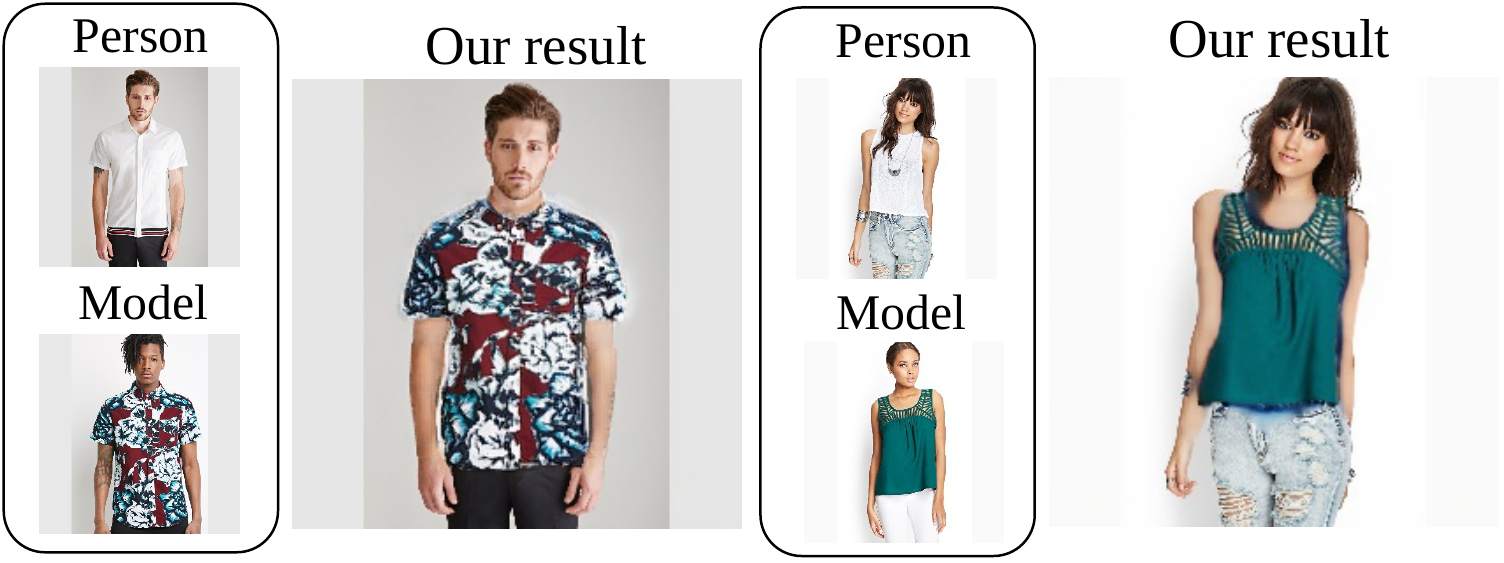}
\captionof{figure}{Illustrates the objective of the present work. Given model and person images, the proposed method generates the image of the person wearing the clothing of the model.}
\label{fig: demo_VTON}
\end{figure}

The VTON problem can be addressed in various settings. Recent VTON methods~\cite{cagan, viton, cpvton, fitme, VTONnew, fashionon, coherence, vtnfp, multiposevton, fwgan, cvpr_2020_2, spvton} require the image of the clothing, in addition to the image of the model (source) wearing that clothing and the person (target). While having a separate clothing image makes the modeling of VTON systems easier, but separate clothes images are rarely available. Shopping websites mostly display images of models posing with the clothes, rather than separate clothes images. Also, people tend to post images on social media platforms donning a variety of clothing items. Hence, the assumption of the availability of a separate clothing image is hard to satisfy in practice; but relaxing this constraint makes the problem more challenging. Because, if a separate clothing image is not available, clothing information has to be extracted from the image of the model. The VTON method we have proposed here works under this relaxed setting, i.e., this method generates the rendition of a customer wearing the clothing of a model when only the model image and the customer image are available (Fig.~\ref{fig: demo_VTON}).

\begin{figure}[h]
\centering
\includegraphics[width=0.65\textwidth]{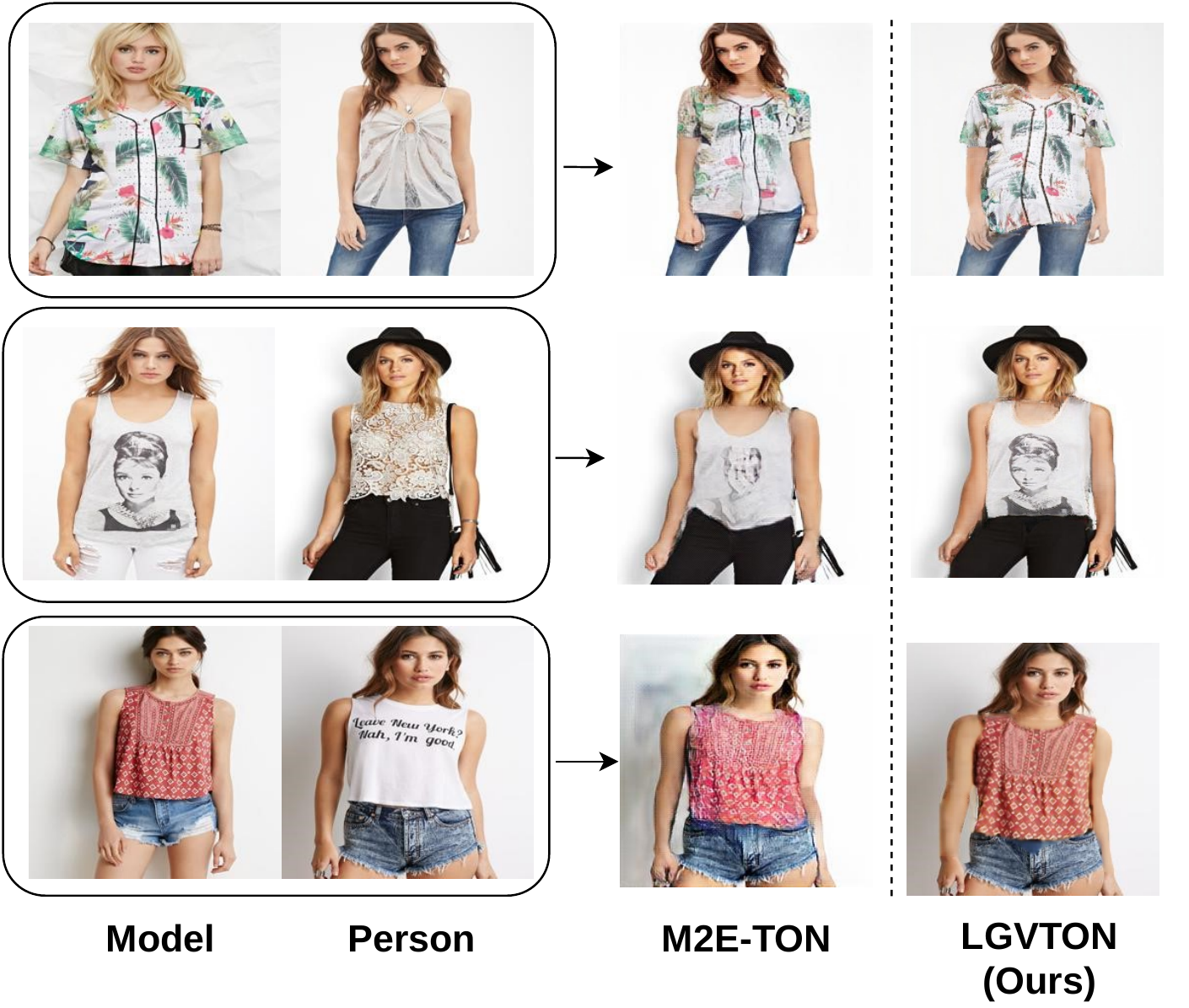}
\caption{Visual comparison with M2E-TON.}
\label{fig: intro_compare}
\end{figure}

Previously, M2E-TON~\cite{m2e} has tried to address the VTON problem where a separate clothing image is not available. However, there are certain drawbacks to that method. A careful analysis of the results (Fig.~\ref{fig: intro_compare}) reveals that it falls short in preserving texture and color in the output (1st and 3rd row of Fig.~\ref{fig: intro_compare}). Besides, the method gets confused with the head pattern printed on clothes and the actual head of a human (2nd row). Another drawback is observed in terms of transferring the true fitting of the cloth. For example, in the 1st row, the sleeves of the shirt are loosely fitted to the model while in the result those are tightly fitted. There are also discrepancies observed in the background of the source and the resultant image.

\begin{figure}[tb]
\centering
\includegraphics[width=0.8\textwidth]{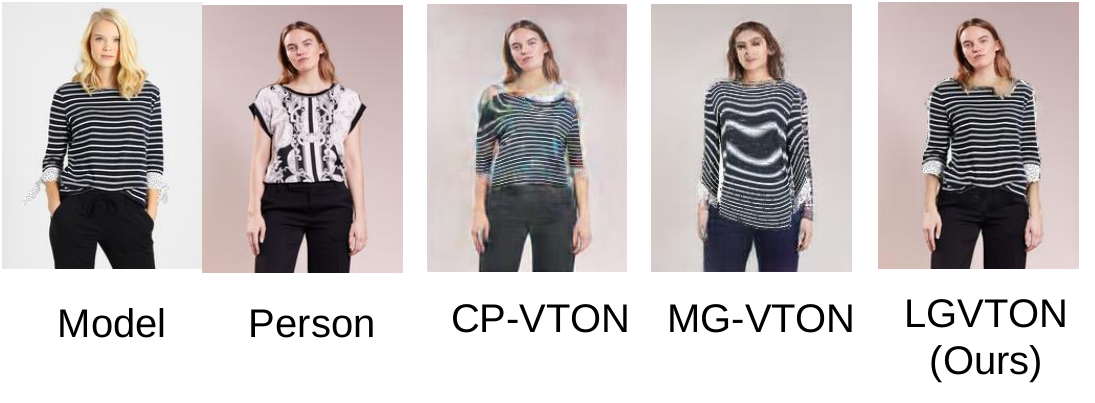}
\caption{Demonstration of issues with geometric matching.}
\label{fig: gmm_issue}
\end{figure}

Of late, VTON methods are following a two-staged  approach~\cite{viton,cpvton,vtnfp,multiposevton}. In the first stage, the clothes are aligned according to the target person's body shape and pose, and, in the second stage, the warped clothes are combined with the target person image. The use of Thin-Plate Spline~(TPS) transformation~\cite{approximatetps1} is popular among VTON methods to warp the source clothes to the target shape. The parameters of this transformation are computed either by shape context warp~\cite{viton} or CNN-based feature matching method called Geometric Matching (GM)~\cite{cpvton, vtnfp, multiposevton}. While the concept of GM was first proposed in~\cite{gmm_source}, CP-VTON~\cite{cpvton} showed its applicability in warping clothes for virtual try-on. Although GM is a popular warping strategy, we observe in Fig.~\ref{fig: gmm_issue} that it results in inaccurate clothing deformations in the presence of complex patterns \cite{laviton,wuton,cvpr_2020_2} including stripes and floral. A possible reason for it might be the substantial flexibility in terms of the TPS transformation realized through the geometric matching network.

\begin{figure}[tb]
\centering
\begin{subfigure}[t]{0.3\textwidth}
    \centering
   	\includegraphics[scale=0.3]{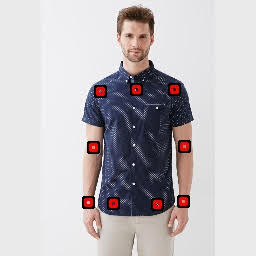}
	\caption{Human landmarks}
\end{subfigure}
\centering
\begin{subfigure}[t]{0.3\textwidth}
\centering
	\includegraphics[scale=0.3]{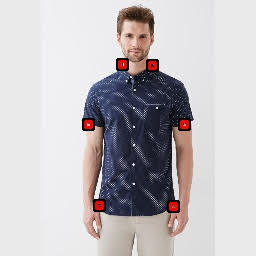}
	\caption{Fashion landmarks}
\end{subfigure}
\caption{Demonstration of two types of landmarks used by LGVTON.}
\label{fig: hlm_flm_demo}
\end{figure}

To address these issues, we propose a novel VTON system, named~\emph{LGVTON} (A Landmark Guided Approach for Model to Person VTON). LGVTON works in three stages using the following modules. \textit{Pose Guided Warping Module (PGWM)} - aligns the model clothes to the body shape and pose of the person. Inspired by the concept of landmark-based image registration~\cite{image_registration}, this module uses the correspondences between two sets of control points (we use human landmarks and fashion landmarks as control points) for warping (Fig.~\ref{fig: hlm_flm_demo}). We show that compared to the existing warping methods, this method works better in the presence of complex patterns in clothes (refer to Fig.~\ref{fig: gmm_issue}). Since human landmarks present the pose of a person, we call our strategy of warping a pose-guided warping. However, the warping generated by PGWM may not be precise near the edges as we are working with only a few control points; this is addressed by the next two modules: Mask Generator Module (MGM) and Image Synthesizer Module (ISM). 
\textit{Mask Generator Module (MGM)}  - attempts to predict the segmentation mask corresponding to model clothing on the person, which aids the next module towards refining the fit of the warped clothing in the VTON output.
\textit{Image Synthesizer Module (ISM)} - combines the warped clothing with the person image to generate the final VTON output. ISM generates a combined mask that aids in preserving the characteristics of the warped clothing and the person image. As a result, this improves the quality of the VTON output.
In order to alleviate the problem of paired data in the `model to person VTON', our method proposes a self-supervised training strategy, and also it does not require any separate clothing image for its execution.

In brief, we make the following contributions. 
\begin{enumerate}
    \item We propose the use of fashion and human landmarks as control points for estimating the warp of the model clothing that should fit onto the target person. We also show its effectiveness in preserving the clothing details after warping.
    \item Unlike most of the methods \cite{viton, cpvton, multiposevton, vtnfp}, our proposed method does not require separate clothes images. 
    Thus it addresses a more practical problem, considering the fact of the rare availability of separate clothes images.
    \item Our image synthesizer generates the final virtual try-on result in a way that not only retains the characteristics of the clothes but also preserves the details of other parts of the person very well. This addresses the problem of loss of clothing details of the target person unlike the results of many existing methods~\cite{cpvton, viton, m2e}.
    \item We propose a self-supervised training strategy that removes the requirement of paired data (model and person wearing the same clothing) for training, which is often difficult to get. 
\end{enumerate}
It is observed experimentally that LGVTON achieves promising performance both qualitatively and quantitatively. We organize this paper in the following way. We discuss an elaborated literature survey in Section.~\ref{related_works}; thereafter we present the methodology of this work in Sec.~\ref{methodology}. A detailed discussion about the training details related to self-supervision is given in Section~\ref{training_details}. In Section.~\ref{experiments} we have presented the qualitative and quantitative experiments, along with an ablation study and human study. The ablation study analyzes the effectiveness of each of the modules of LGVTON. Similar to~\cite{viton,cpvton}, LGVTON focuses on upper body cloths only. However, this is due to the limited number of available fashion landmarks for lower body cloths. Whereas, it is worth mentioning that our method can be extended for lower body clothes in the presence of a sufficient number of landmarks. This paper contains an appendix, which provides useful details related to this work along with more results of LGVTON on different datasets.
\section{Related Works}
\label{related_works} 
In this section, the existing works related to VTON are briefly reviewed. We also discuss relevant studies on pose estimation, human parsing methods which are used in our work as preprocessing steps.
\subsection{Human parsing and pose understanding}
\label{sec: related_works_pose} 
Pose estimation is the method of localizing anatomical key points of human, called \textit{human landmarks}. Human parsing, on the other hand, refers to the segmentation of the human image into multiple parts with fine-grained semantics. Researchers have used these techniques in many tasks, including human behavior analysis \cite{mabrouk2018abnormal} and person re-identification \cite{zhao2013unsupervised}. Cao et al.~\cite{openpose} proposed a Part Affinity Field (PAF) based method for localizing human landmarks. PAF is a non-parametric representation that learns to associate the body parts of the person in the image. We use the method presented in \cite{openpose} for localizing the human landmarks of the model and the person. We use these landmarks in computing the target warp of the model clothes that fit the target person. However, aligning the model clothes requires extracting the clothes first. We use the human parsing network in \cite{lipssl} for this purpose. It uses a self-supervised structure-sensitive learning framework enforcing the consistency between parsing results and the human joint structures. The basic idea of VTON methods not only involves computing a precise target warp but also synthesizing a realistic try-on output by integrating the target warp with the target person image. However, this is invariably related to a better reconstruction of the human body under the clothing of the person, as removing the old clothes and putting on the new ones might generate the need of reconstructing some previously covered body parts. This is satisfied in the proposed method by providing the densepose~\cite{densepose} representation of the person as input to our ISM module. A densepose~\cite{densepose} representation maps all human pixels of an RGB image, to the 3D surface of the human body, thus providing a precise estimate of the human body shape under the clothing.
\subsection{Visual fashion analysis}
Visual fashion analysis has attracted wide attention over the last few years. Localizing fashion landmarks~\cite{flmwild}, \cite{wang2018attentive}, clothing category classification~\cite{wang2018attentive} are some of the key issues addressed in this domain. Fashion landmarks contain information about the structure of the clothes~\cite{flmwild, wang2018attentive} which may be useful for VTON. However, to the best of our knowledge, the importance of these landmarks has not been explored rigorously in the context of VTON. This work shows experimentally that these landmarks may play a significant role in predicting better target warp of the model clothes.
\subsection{Virtual Try-on (VTON)} 
Virtual Try-On (VTON) systems based on the 3D body shape of the person~\cite{pons2017clothcap}, \cite{sekine2014virtual} produce enhanced VTON experience but incur a significant cost of computation and require expensive infrastructure to capture the 3D data. 
To address these issues, recent methods of VTON~\cite{cagan, viton, cpvton, fitme, VTONnew, fashionon, coherence, m2e, vtnfp, multiposevton, fwgan, cvpr_2020_1, cvpr_2020_2, sievenet, clothvton, han2019clothflow, 2018_CVPR_apptrans} have explored the image-based approach to solve the problem, which is less resource-intensive and has a lot of scopes as well. However, generating a perceptually convincing output without 3D data of the clothes and the person is quite challenging.

The problem of image-based virtual try-on has gained immense interest in the last few years. Various new interesting ideas have evolved. For example, in Swapnet~\cite{swapnet} the authors propose a Conditional Generative Adversarial Network (cGAN)~\cite{cgan} based two-stage pipeline to swap clothing between a pair of person images. Mihai et al.~\cite{2018_CVPR_apptrans} proposed a method that transfers appearance from a source person image onto the target person image while preserving the clothing segmentation layout of the source person. Related to unpaired data, a new method has been proposed~\cite{cvpr_2020_1} that allows visualizing the composition of clothing items, selected from various reference images, on a person.

The objective of appearance transfer methods is to transfer the complete appearance. However, the goal of this paper is to transfer the desired clothing only. This problem has been addressed previously first by CAGAN~\cite{cagan} which proposed a cGAN trained with cycle consistency loss to learn the relation between clothes and their appearance rendered on a human. But its training requires images of the same person wearing different clothes in the same pose, which is uncommon. This is addressed by VITON~\cite{viton} using a coarse-to-fine framework for VTON. Its warping module uses shape context warp~\cite{shape_context} which is a hand-crafted feature-based point correspondence warping strategy. The other two modules combine the clothes and the person. It is observed that VITON often fails to keep the details of the warped clothes which is addressed by CP-VTON~\cite{cpvton}. This method predicts a mask that helps to retain the characteristics of the clothes. It employs the concept of geometric matching (GM)~\cite{gmm_source} in the domain of clothes, but due to its high flexibility, it often causes undesirable warping results in the presence of complex patterns in clothes. While CP-VTON and VITON focus more on the warped clothes details, they often generate inaccurate bottom clothes and imprecise body part details of the person. VTNFP~\cite{vtnfp} addresses this issue, thereby generating better results. However blurry clothes details, body parts, and artifacts remain to some extent in its results. Another method SP-VTON addresses the incorrect body shape reconstruction problem of CP-VTON and VITON employing the densepose~\cite{densepose} representation to estimate better 2D body shape under clothes. However, as observed from its results, it falls short in terms of reconstructing accurate clothes details. A recently proposed method ACGPN~\cite{cvpr_2020_2} uses a semantic layout-based approach to VTON and shows an improvement over VTNFP in terms of perceptual quality of results. Another novel approach proposed by MG-VTON~\cite{multiposevton} addresses the problem of multi-pose VTON. It relaxes the pose constraint in VTON problems. However, it is observed that the details of persons are not well preserved in its results. Some other multi-pose methods are~\cite{fitme, zheng2019virtually, fashionon}.

Although the above methods have contributed significantly in the domain of VTON, these require a separate clothes image as input, therefore, imposing a constraint in terms of data availability. In this regard, M2E-TON~\cite{m2e} proposed a method to transfer clothing from model to person, eradicating the necessity of separate clothing images. It aligns the model image to the shape and pose of the target person using their dense pose~\cite{densepose} representations, followed by refining the textures of the clothes image and finally merging with the person image using a fitting network. Another method~\cite{zeng2020tilegan} proposed a two-stage image generation approach for generating clothing images from model images. The authors used the implementation of CP-VTON to show that with their reconstructed clothes image 'a model to person try-on problem' can be reduced to 'clothes to person try-on problem'. However, this way of solving VTON may not be economical due to the additional stages for clothes reconstruction.

Compared to the above methods, we propose a novel method of warping that considers human and fashion landmark correspondences for computing the target warp. We also propose an image synthesizer module that produces an enhanced VTON output retaining all necessary details of the clothes and the person. Similar to M2E-TON, our method also addresses the model-to-person VTON scenario, which is more pragmatic.

Other than image-based VTON, at present, the research community is also focusing on video-based VTON~\cite{fwgan}, which is beyond the scope of the present discussion. 
%
\section{Proposed method}
\label{methodology} 
\begin{figure*}[h]
	\centering
	\includegraphics[scale=0.61]{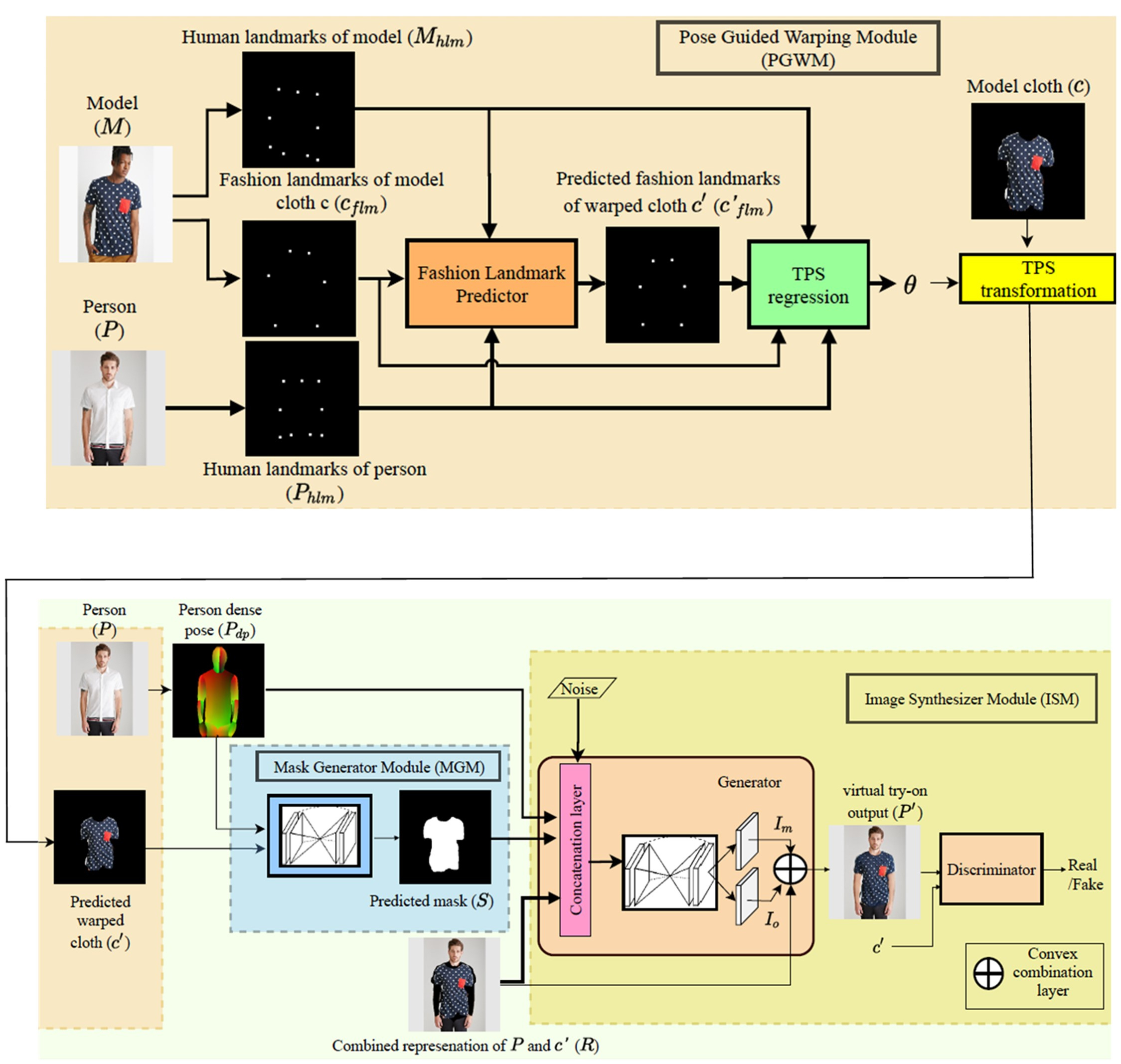}
	\caption{Block diagram depicting workflow of LGVTON.}
	\label{fig: final_blockdia}
\end{figure*}

We propose a Landmark Guided Approach for Virtual Try-On (LGVTON) that learns to synthesize an image of a person wearing a model's clothing. Formally, given a model image $M$ wearing the clothes $c$ and a person image $P$, LGVTON synthesizes $P'$, which is the new image of the person wearing the model clothes.
The workflow of LGVTON is three-fold (refer to Fig.~\ref{fig: final_blockdia}). First, it attempts to warp the model clothes according to the shape and pose of the target person. This is done by our pose guided warping Module (PGWM). Second, a segmentation mask corresponding to the clothing area of the person wearing the target clothing is predicted by our mask generator module (MGM). Third, the image synthesizer module (ISM) synthesizes the final virtual try-on output.

We first mention various notations used in this paper. As mentioned, $M$ and $P$ denote the model image and the person image respectively. $P'$ is the person image after virtual try-on (our result). $c$ and $c'$ denote the model clothes and target warp of the model clothes, respectively. We assume that each of them is an RGB image with the same dimension. We represent a landmark as a 2-tuple $\in \mathbb{N}^{2}$.
We denote the multiset of human landmarks of the person $P$ by $P_{hlm} = \{\boldsymbol\alpha^P_1, \boldsymbol\alpha^P_2, \ldots,
\boldsymbol\alpha^P_{n_h}\}$ and that of the model $M$ by $M_{hlm} = \{\boldsymbol\alpha^M_1, \boldsymbol\alpha^M_2, \ldots,
\boldsymbol\alpha^M_{n_h}\}$, where $n_h$
is the total number of landmarks of a human used in this work. We further denote the multiset of fashion landmarks of the clothes of the model by $c_{flm} = \{\boldsymbol\beta^c_1, \boldsymbol\beta^c_2, \ldots,
\boldsymbol\beta^c_{n_f}\}$ and that of the target warp of the model clothes by $c'_{flm} = \{\boldsymbol\beta^{c'}_1, \boldsymbol\beta^{c'}_2, \ldots, \boldsymbol\beta^{c'}_{n_f}\}$, where $n_f$ is the number of fashion landmarks of a clothing used in this work. $\mathcal{F}$
denotes the fashion landmark predictor network in PGWM, which is a learnable function. $P_{dp}$ and $M_{dp}$ denote the Densepose~\cite{densepose}
representations of the person and the model, respectively. $S$ is the target mask predicted by our MGM and $R$ is the combined representation of $P$ and $c'$ (an input to ISM).

\subsection{Computing the target warp of the model clothing} 
This section elaborates on our Pose Guided Warping Module (PGWM). It computes a thin-plate spline (TPS)~\cite{duchon} transform $f(\cdot)$ to warp $c$ into $c'$, where $c'$ is aligned with the the body shape and pose of the person $P$. $c$ is obtained from $M$ using the human parsing network proposed in~\cite{lipssl}. Our method of predicting the target warp is inspired by the concept of landmark-based image registration~\cite{image_registration}, which uses landmark (also referred to as control points) correspondences between two images to estimate the parameters of the transformation from one image to the other. Since landmarks define the structural information, therefore using these points allows the transformation to be interpreted in terms of the underlying anatomy. In our case, we consider fashion and human landmarks (Fig.~\ref{fig: hlm_flm_demo}) as control points. Once the parameters of the transformation are computed, the target warp $c'$ is computed by applying the corresponding TPS transformation on $c$. Therefore, the generation of $c'$ involves two steps discussed below.

\subsubsection{Estimating the source and the target sets of landmarks} 

We extract human landmarks $M_{hlm}$ and $P_{hlm}$ corresponding to model $M$ and person $P$, respectively, using the method proposed by Cao et al.~\cite{openpose}. These landmarks are anatomical key points of human that are generally used to represent the pose of a person. However, as Bogo et al.~\cite{smplify} suggested, these landmarks can also approximate the body shape of a person. Hence, the correspondence between the human landmarks of the model and the person can represent the structural change of $c$ to $c'$.

\begin{figure}[h]
	\centering
	\includegraphics[width=0.95\linewidth]{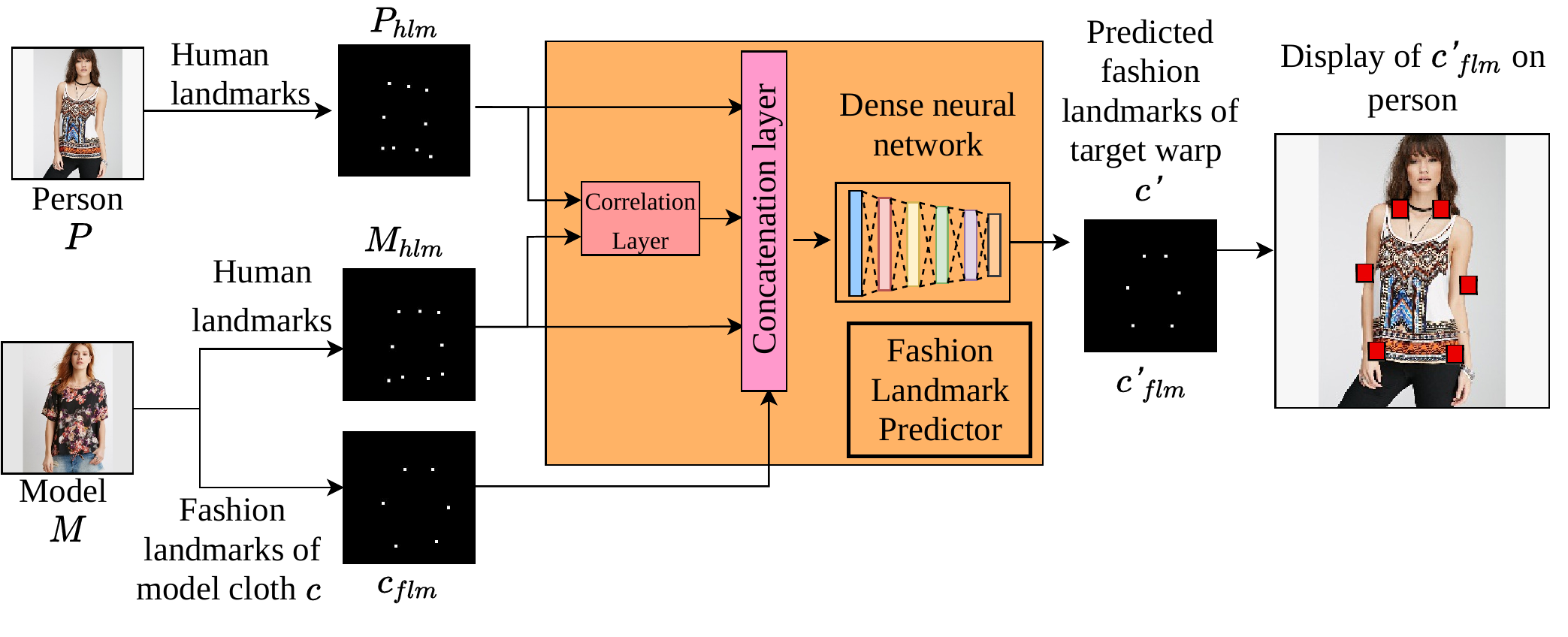}
	\caption{Block diagram of fashion landmark predictor network ($\mathcal{F}$) in Pose guided warping module (PGWM).}
	\label{fig: pgwm1}
\end{figure}

However, computing the deformation of a non-rigid object such as clothes is very challenging since it includes the deformation of designs or patterns present in the clothes. Achieving a highly accurate warping, which is very essential for realistic VTON output, usually requires many landmarks. Consequently, along with human landmarks, we consider fashion landmarks as well so that the warping becomes more precise.\footnote{\label{ablation_supplementary}This is elaborated more in the ablation study given in the Appendix.}
Since each image in DeepFashion~\cite{deepfashion} dataset is annotated with fashion landmarks, so we have ground-truth for fashion landmarks of $c$. However, that of $c'$ is not available. Hence, to compute corresponding landmarks of $c'$ we propose a fashion landmark predictor network $\mathcal{F}$ (Fig.~\ref{fig: pgwm1}) that predicts ${c'}_{flm}$ given ${M}_{hlm}$, ${P}_{hlm}$ and ${c}_{flm}$.
This network is trained using L2 loss.

Here, we are transferring the clothes from model to person, hence, it has to undergo deformation according to the way the body shape and pose changes from model to person. We incorporate this observation in $\mathcal{F}$ by introducing a correlation layer with ${M}_{hlm}$ and ${P}_{hlm}$ as inputs, in view of the fact that statistical correlation represents the linear relations between a pair of variables. An ablation study related to this layer is given in Section~\ref{ablation}\footnote{\label{correl_foot}A more detailed discussion on the correlation layer is given in the Appendix.}.

Training this network is tricky since paired data, i.e., human and fashion landmarks of two persons wearing the same clothes in any arbitrary pose, is not available. We discuss more on this in Section~\ref{training_details}.
\subsubsection{Computing the target warp}
Once we have the source landmarks ${M}_{hlm}$, ${c}_{flm}$, and the target landmarks ${P}_{hlm}$, ${c'}_{flm}$, we now define $\mathcal{R} = \{\boldsymbol\alpha^M_1, \boldsymbol\alpha^M_2, \ldots, \boldsymbol\alpha^M_{n_h}, \boldsymbol\beta^c_1, \boldsymbol\beta^c_2, \ldots, \boldsymbol\beta^c_{n_f}\}$ and $\mathcal{T} = \{\boldsymbol\alpha^P_1, \boldsymbol\alpha^P_2, \ldots, \boldsymbol\alpha^P_{n_h}, \boldsymbol\beta^{c'}_1, \boldsymbol\beta^{c'}_2, \ldots, \boldsymbol\beta^{c'}_{n_f}\}$ (refer to the description of these notations given at beginning of Section~\ref{methodology}). Our idea is to utilise the correspondences between the source $\mathcal{R}$ and the target landmarks $\mathcal{T}$ to transform $c$ to $c'$. Hence, we find a smooth interpolation function
$f: \mathbb{N}^{2} \rightarrow \mathbb{N}^{2}$, such that, $f(\boldsymbol\alpha^M_i) = \boldsymbol\alpha^P_i;~ i = 1, 2, \ldots, n_h;$ and $f(\boldsymbol\beta^c_i) = \boldsymbol\beta^{c'}_i;~ i = 1, 2, \ldots, n_f;$ hold. Then we employ $f(\cdot)$ to transform $c$ to $c'$; which means basically $f(\cdot)$ is applied on the mesh grid containing $c$ to get $c'$. For notational convenience, we use $\mathbf{r}_j$ and $\mathbf{t}_j$ to denote the $j^{th}$ element of $\mathcal{R}$ and $\mathcal{T}$ respectively, where $j = 1,2, \ldots, N$. Here $N$ denotes the total number of elements in each of the multiset $\mathcal{R}$ and $\mathcal{T}$.

Now, the deformation of non-rigid objects such as clothes should be smooth. Considering this, we choose $f(\cdot)$ to be a thin-plate spline (TPS) transform~\cite{duchon}, which is a widely used transform representing coordinate mappings with a penalty term for imposing smoothness~\footnote{\label{tps_foot}a detailed study on TPS is given in the Appendix.}. As we are dealing with only estimates of the true landmark locations which may be noisy, so instead of exact interpolation in this step, we remain content with approximation. This is accomplished in TPS transform by minimizing the following objective function~\cite{approximatetps1, approximatetps2},
\begin{equation}
	H[f] = \sum_{j=1}^N \|f(\mathbf{r}_j)-\mathbf{t}_j\|_2^2 + \lambda \iint\limits_{\mathbb{R}^2} [f_{xx}^2 + 2f_{xy}^2 + f_{yy}^2]dx\ dy,
\end{equation}
where $\lambda$ is a regularization parameter, which determines the relative weight between the approximation behavior and the smoothness of the transformation. It is a positive scalar. $f_{xx}$, $f_{xy}$, $f_{yy}$ are second-order gradients of $f(\cdot)$ as we consider each landmark as a 2-tuple.

\begin{figure}[h]
	\centering
	\includegraphics[scale=0.65]{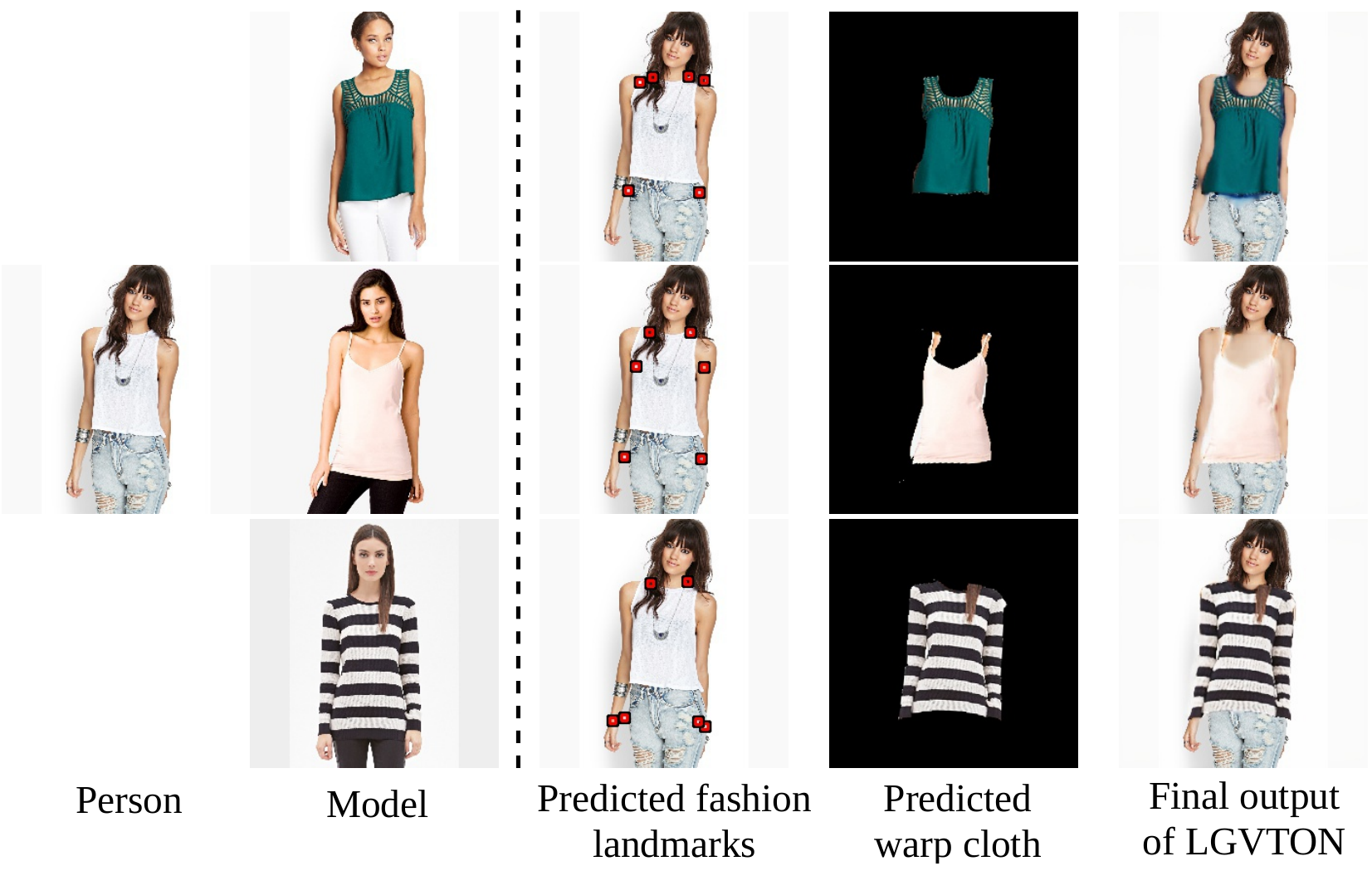}
	\caption{Results of the two steps of PGWM and final output of LGVTON.}
	\label{fig: lgvton_allsteps}
\end{figure}

Note that, this work is targeted to upper body clothes only, so for warping, we use only upper body human landmarks which in total is 9, and fashion landmarks corresponding to upper body cloths which in total is 6 (refer to Fig.~\ref{fig: hlm_flm_demo}). Therefore, we have $N$ = 15 landmarks, i.e., $n_h$ = 9 and $n_f$ = 6. We select $\lambda$ = 0.01, experimentally for satisfactory results. Some target warps generated by PGWM are shown in Fig.~\ref{fig: lgvton_allsteps}.
%
\subsection{Generating Segmentation Mask of Target Clothing}
\begin{figure}[tb]
	\centering
	\includegraphics[scale=0.65]{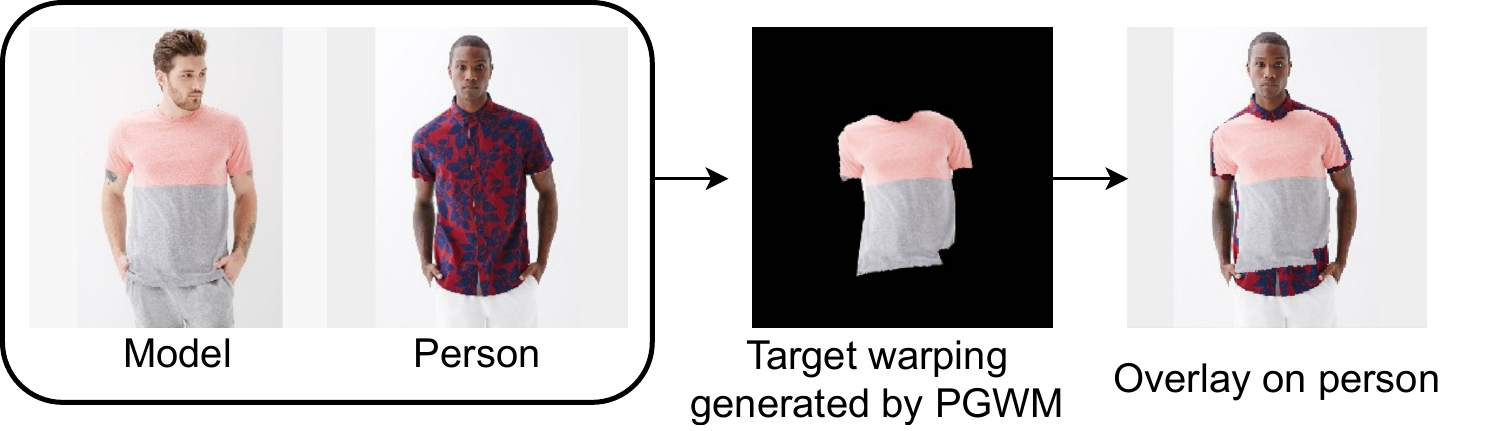}
	\caption{Demonstration of warping glitch. The target warping generated by the PGWM is not completely accurate as it is observable when overlaid on the person image. Although the overall fit is good the fit is not precise near the areas of the collar and sleeves.}
	\label{fig: warp_problem}
\end{figure}

Human landmarks are a sufficient representation of pose and a good approximator of body shape~\cite{smplify} but not a highly accurate estimator~\cite{shigeki2018estimating}. That means they result in a good overall warping but it might not be very precise near the edges of the cloth, as illustrated in Fig.~\ref{fig: warp_problem}. We address this problem as a warping glitch. This is addressed by our Mask Generator Module (MGM). MGM predicts the segmentation mask corresponding to the region of the clothes of the target person after wearing the model clothing. This in turn guides the next module (ISM) to handle the warping glitches.

Formally, given $c'$ and densepose~\cite{densepose} $P_{dp}$ of the person (refer to Fig.~\ref{fig: final_blockdia}), MGM generates the target clothing segmentation mask $S$ corresponding to $c'$ (ideally in cases of no warping glitch, mask of $c'$ should be same as $S$). Densepose contains 24 part labels of human, where each part has UV parametric values of the body surface~\footnote{\label{supp_mgm_abl}A detailed ablation study of MGM is given in the Appendix.}. Similar to the previous module, paired data is also not available for training this module. However, we tackle this using a landmark perturbation based approach, discussed in Section~\ref{training_details}. 

\subsection{Synthesizing VTON image} 

Our final module, the \textit{image synthesizer module} (ISM) generate the final virtual try-on output $P'$ from the inputs $c'$ and $P$.

This module takes in the following inputs: (i)~The generated target clothes mask $S$ from MGM, (ii)~a combined representation $R$ of $P$ and $c'$, Densepose $P_{dp}$ of $P$. Note that instead of providing $P$ and $c'$ separately, we provide a combined representation $R$ as input as it enhances the perceptual quality of the output. This is discussed more in detail later in this section. This representation is obtained by setting pixel values in the upper body area including the clothing region of $P$ to zero (using the human parsing of $P$) and then combining $c'$ with it. Such a representation can be viewed in the block diagram given in Fig.~\ref{fig: final_blockdia}. This module is implemented as a conditional generative adversarial network (cGAN). Therefore, an additional input which is the random noise $\textbf{z}$ sampled from a noise distribution $p_\textbf{z}$ is provided.

ISM when trained without $S$ generates image artifacts. This is due to the great variety in the types of designs of the clothes, which confuses the cGAN to distinguish a warping glitch from a clothing design. Therefore, the network can not identify a warping glitch itself. Whereas, giving $S$ as an input makes the network identify the regions of warping glitches. For a more detailed discussion on this please see supplementary material.

The objective of our cGAN may be expressed as 
\begin{flalign}
	L_{\text{cGAN}}(G, D)  = \mathbb{E}_{\mathbf{x} \sim p_{data}} [\log D(\mathbf{x}|\mathbf{y})] 
	+ \mathbb{E}_{\mathbf{z} \sim p_{\mathbf{z}}} [\log(1 − D(G(\mathbf{z}|\mathbf{y})))], 
\end{flalign}
where the generator $G$ learns a distribution $p_g$ over data $\textbf{x}$. It builds a mapping function from a prior noise distribution $p_\textbf{z}$ with a conditional information $\textbf{y}$ to the data space. While the discriminator $D$ represents the probability of $\textbf{x}$ given $\textbf{y}$, to be coming from the training data rather than the generator distribution $p_g$~\cite{cgan}. cGAN is trained to minimize an objective $L_{\text{cGAN}}$ against an adversarial D that tries to maximize it. The optimum $G$ denoted by $G^*$ is 
\begin{equation}
	G^* =~ \text{argmin}_G~ \text{argmax}_D~ L_{\text{cGAN}} (G, D).    
\end{equation}
In our model, $c'$ is the condition given to both $G$ and $D$.

$G$ contains an hourglass network~\cite{hourglass} (a convolutional neural network with skip connections~\footnote{\label{supp1}more detailed in the Appendix.}), followed by two parallel convolution layers, giving activation $I_o$, an intermediate VTON output, and $I_m$, a mask, which helps to retain the necessary details from $R$. The last layer of $G$ is a convex
combination layer that combines $I_o$ and $R$ using $I_m$. Generally, the averaging tendency of convolution operation causes the loss of fine details in the output. The network tackles this with the help of $I_m$ as it aids in preserving necessary details from $R$ in the final output. This can be
validated from our results presented in Sec.~\ref{experiments} which shows that our result retains cloth, person, and background details better in comparison to the results of the other methods. The discriminator $D$ is a patchGAN discriminator~\cite{pix2pix}.

It has been experimentally observed by different past works on conditional GAN \cite{perceptual_loss},\cite{textureGAN} that having other loss functions, such as perceptual loss~\cite{textureGAN}, along with adversarial loss gives a better output. Therefore, we incorporate structural dissimilarity index (DSSIM) and VGG perceptual loss~\cite{perceptual_loss} as additional loss functions in the generator. The inclusion of these additional loss terms keeps the task of the discriminator unchanged while the generator, in addition to the task of fooling the discriminator, has to generate data instances closer (in L2 sense) to the ground-truth.

SSIM (Structural similarity index)~\cite{ssim} is an image metric that measures the structural similarity between the two images. However, in the neural network, the objective is to minimize the value of the loss function, so instead of SSIM we take DSSIM that is related to SSIM the following way, $\text{DSSIM} (\cdot, \cdot) = 1 - \text{SSIM} (\cdot, \cdot)$. 
DSSIM loss for the generator is defined as follows 
\begin{equation}
	L_{\text{DSSIM}}(G) = \mathbb{E}_{\mathbf{z} \sim p_{\mathbf{z}}, \mathbf{x} \sim p_{data}} \text{DSSIM}(\mathbf{x} , G(\mathbf{z}|\mathbf{y})),
\end{equation}
where $\mathbb{E}$ denotes expectation.

VGG perceptual loss~\cite{perceptual_loss} is also a L2 loss between the features of generated and ground-truth images, obtained from different layers of the pretrained classification network (VGG-19). Instead of exactly matching the pixel values of the generated and ground-truth images, this loss matches their feature representations. This encourages the network to produce images that are perceptually similar to their corresponding target images. Formally, this loss is defined as 
\begin{flalign}
	L_{\text{VGG}}(G) = \mathbb{E}_{\mathbf{z} \sim p_{\mathbf{z}}, \mathbf{x} \sim p_{data}} \left[\sum_{i=1}^{\rho} \frac{1}{C_i H_i W_i} \|(F_{i}(\mathbf{x})  -  
	F_{i}(G(\mathbf{z}|\mathbf{y}))\|^2_{2}\right],
\end{flalign}
where $F_{i}(\textbf{x})$ denotes the activation at the $i^{th}$ layer of VGG-19 for the input image $\textbf{x}$. $\rho$ is the total number of layers of VGG-19 that we are using. Physically $F_{i}(\textbf{x})$ is a feature map of shape $C_i \times H_i \times W_i$, where $C_i$, $H_i$, and $W_i$ denote the number of channels, height, and width of the $i^{th}$ feature map respectively. We take the features from conv$1\_2$, conv$2\_2$, conv$3\_2$, conv$4\_3$, conv$5\_1$ layers of VGG-19. 

Hence, our final objective function becomes 
\begin{equation}
	G^{**} = ~\text{argmin}_G~\text{argmax}_D~ (L_{\text{cGAN}}(G, D) +  
	L_{\text{DSSIM}}(G) + L_{\text{VGG}}(G) )
\end{equation}

\section{Training details}
\label{training_details} 

Training of LGVTON is tricky since paired data as shown in Fig.~\ref{fig: ideal_pair} is not usually available in the publicly available datasets~\cite{deepfashion, multiposevton} related to fashion. In this section, we discuss our training strategies for different modules.
\subsection{Pose Guided Warping Module (PGWM)}
This module has one trainable component which is the fashion landmark predictor network $\mathcal{F}$. Given the human landmarks of the model and the person and the fashion landmarks of the clothing of the model, $\mathcal{F}$ predicts the fashion landmarks of the target warp of the model's clothing. While $\mathcal{F}$ deals in landmarks only, but the inputs and the corresponding ground-truths are extracted from the respective model and person images for preparing the training data. Below, while discussing the training data, we refer to the images corresponding to the landmarks instead of the landmarks, for maintaining the simplicity of the explanation.

\begin{figure}[h]
	\centering
	\includegraphics[width=0.8\linewidth]{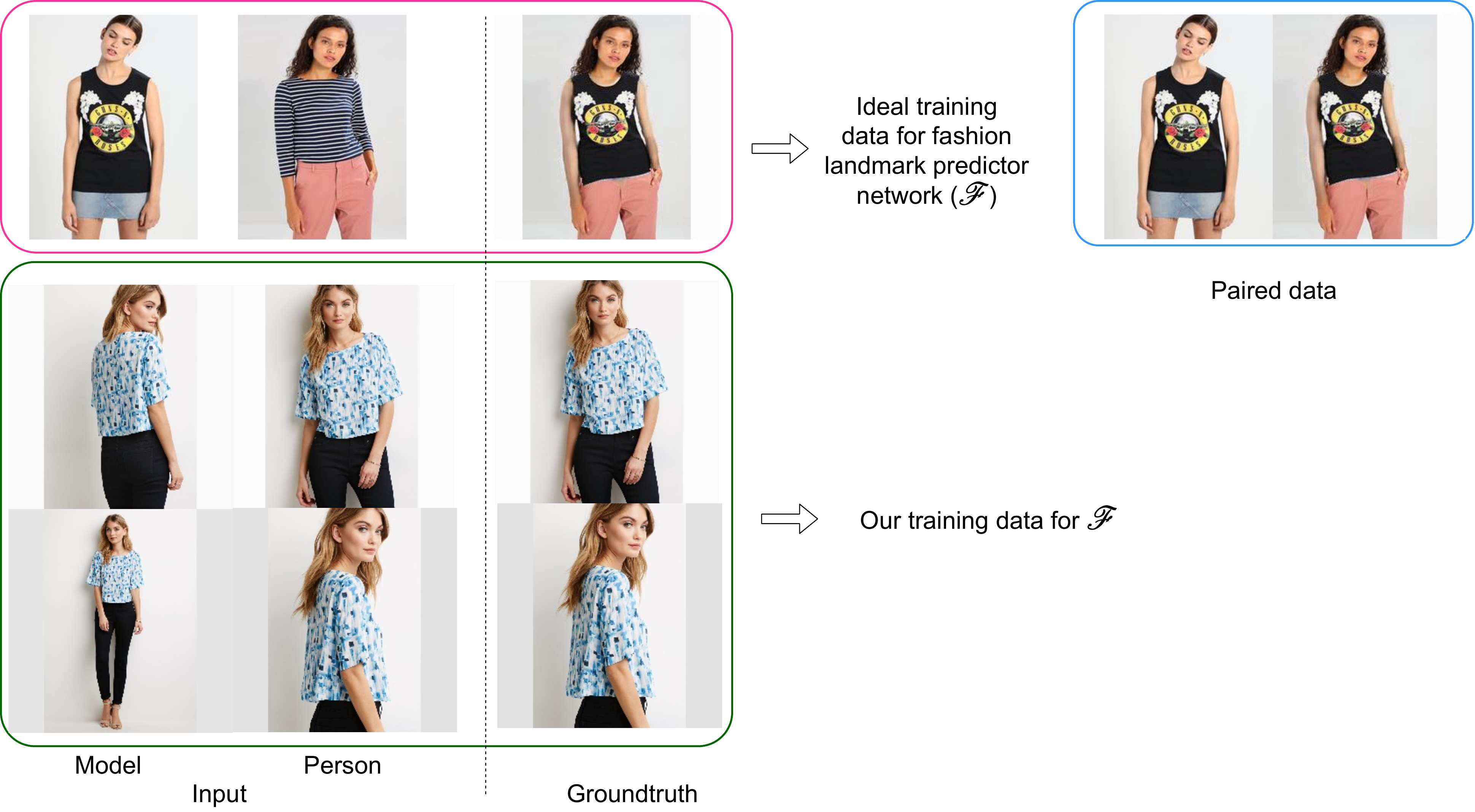}
	\caption{
		Example of an ideal training data of LGVTON and our training data for $\mathcal{F}$. Although $\mathcal{F}$ operates on landmark data only, for keeping the illustration simple we have shown only the images of the models and the persons from which the landmarks are extracted for preparing the training data. Note that since the ground-truth corresponding to the ideal training data does not exist, hence, for illustration, we used the output of LGVTON.}
	\label{fig: ideal_pair}
\end{figure}

We train $\mathcal{F}$ using the data pairs of the same person wearing the same clothing in different poses. In the way we utilize, these data pairs have sufficient variability of poses between the model and the person. Moreover, such data pairs are available in most of the fashion datasets~\cite{deepfashion,multiposevton}. Please see Fig.~\ref{fig: ideal_pair} where we have given an example of our training data as well as an ideal training data for a better understanding of the reader. However, our training data pairs are devoid of basic clothing structure variability, as the model and person are wearing the same cloth. By clothing structure, we mean design style like sleeve length and neck shape, etc., which are supposed to be encoded by fashion landmarks. At this point, we disregard texture and color of clothes as $\mathcal{F}$ deals with landmarks only. We observed that our network generalizes well across different input clothing shapes. Note that we can not measure the performance of this network separately due to the lack of ground-truth data; however, the final try-on outputs in a way reflect the performance of each of its component modules including this network. Some results of $\mathcal{F}$ on different clothing shapes and poses are shown previously in Fig.~\ref{fig: lgvton_allsteps}.
\subsection{Mask Generator Module (MGM)}
\begin{figure}[tb]
	\centering
	\includegraphics[scale=0.4]{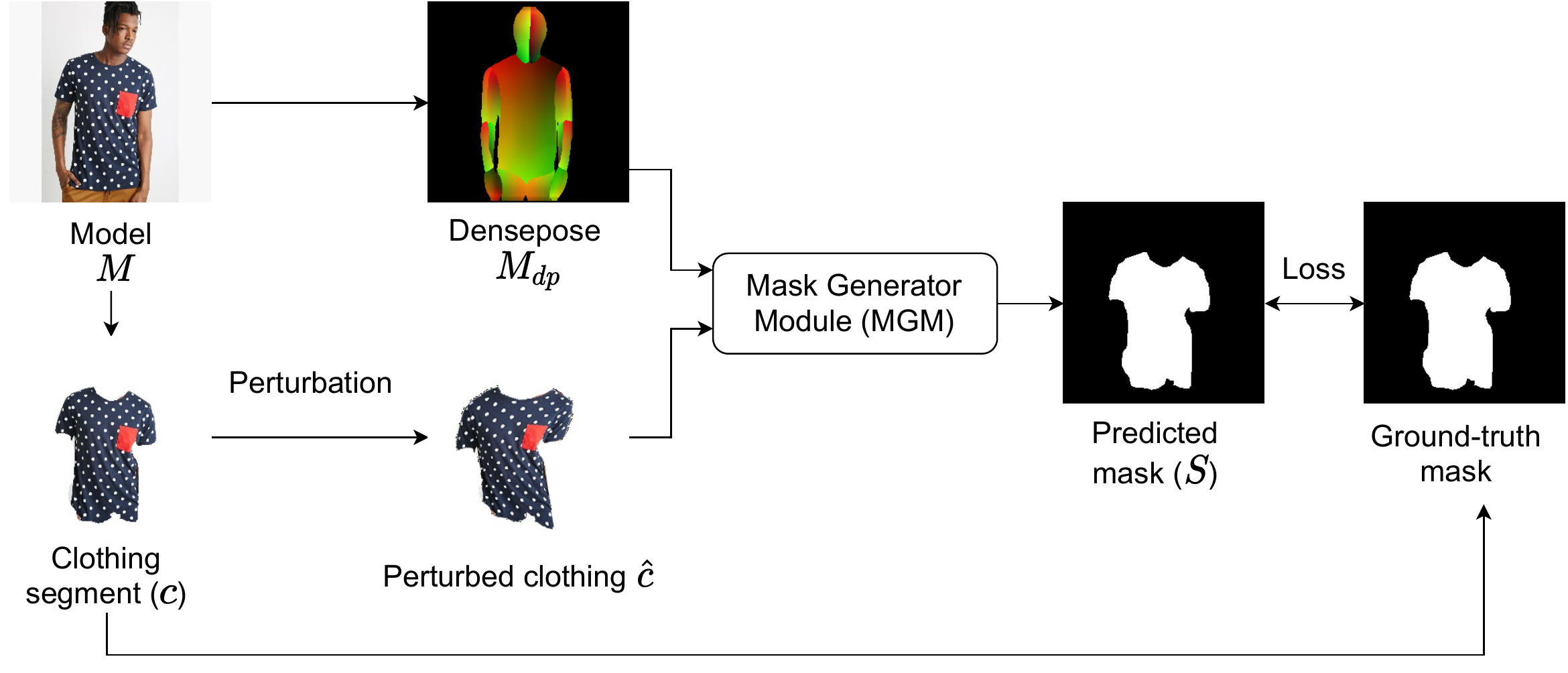}
	\caption{Illustration of the training scenario of mask generator module.}
	\label{fig: mgm}
\end{figure}

Given the target warp along with the densepose of the person, the objective of this module is to predict the mask of the clothing region on the target VTON output. The given target warp (estimated by the PGWM) may contain warping glitches. So the idea is to make the network learn to identify warping glitches and discard their effects in the predicted mask. An example of a training scenario of this mask generator module (MGM) is shown in Fig.~\ref{fig: mgm}. To train this network, for each model image of the training dataset, we extract its clothing region (segment) with the help of a human parsing method~\cite{lipssl} and artificially induce a warping glitch on it. The ground-truth corresponding to this input is the mask of the clothing segment before inducing the warping glitch. An example of the training data of MGM is shown in~Fig.~\ref{fig: mgm}.

To induce a warping glitch artificially on a clothing segment, say, $c$, we compute $\hat{c}$ by random perturbation of ${c}_{flm}$. In this work, we perturb ${c}_{flm}$ by adding random noise $\mathcal{N}(0,0.001)$ to it and denote it by ${\hat{c}}_{flm}$. 
A TPS transformation mapping from ${c}_{flm}$ to ${\hat{c}}_{flm}$ is computed and used to warp $c$ to $\hat{c}$. Multiple such random warps corresponding to each model image and the associated densepose representation of the model are computed to prepare the training data of this module. Training with multiple such perturbed clothing segments corresponding to each clothing segment, make the network learn to extract features required for predicting the desired mask drastically reducing the effect of landmark perturbations, i.e., warping glitches.
\subsection{Image Synthesizer Module (ISM)}
\begin{figure}[!h]
	\centering
	\includegraphics[scale=0.3]{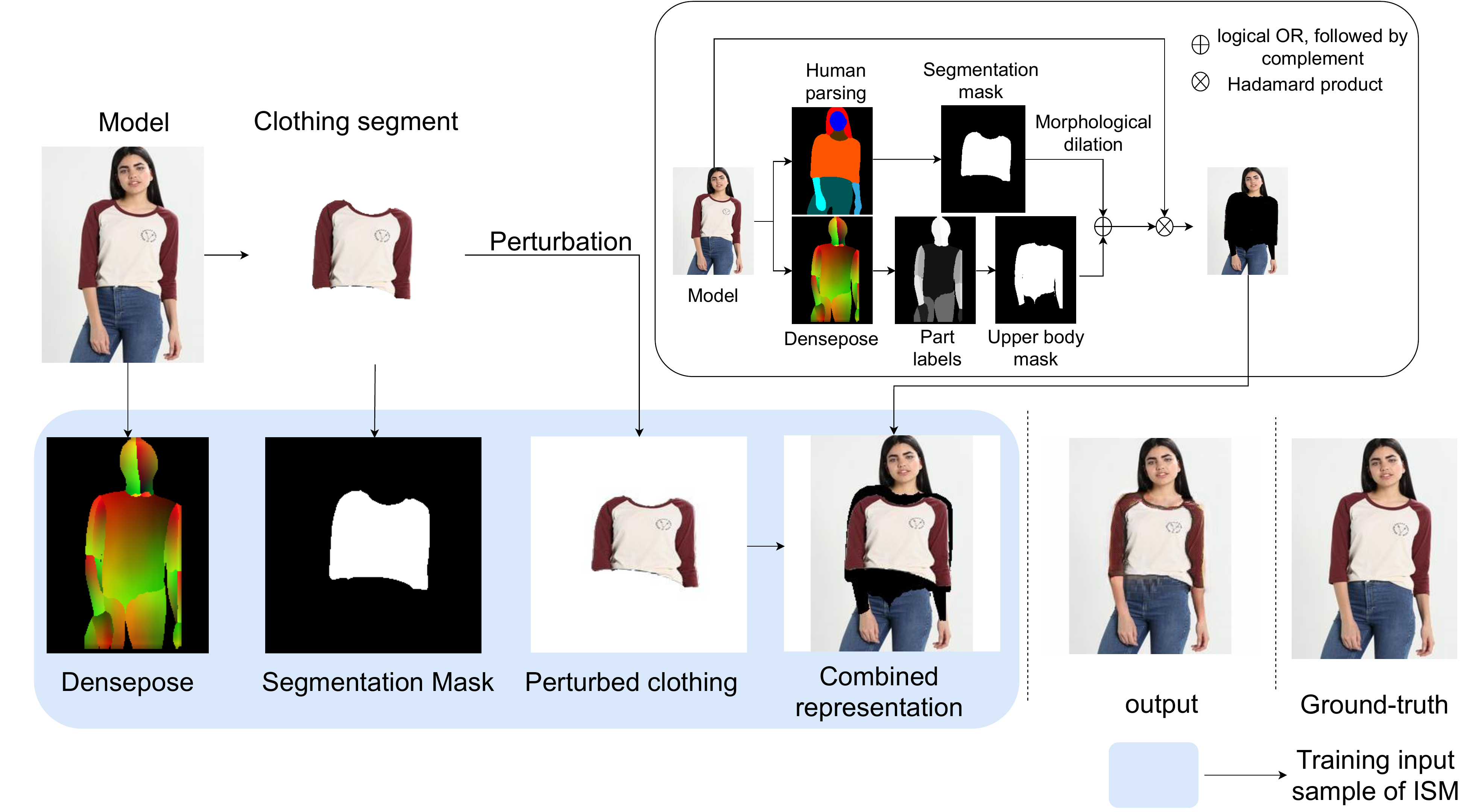}
	\caption{Example of a training data sample of image synthesizer module (ISM).}
	\label{fig: training_data_ism}
\end{figure}

The objective of ISM is to combine the warped clothing with the person image seamlessly. Now for a pair of images of distinct model and person, the ground-truth image of the person wearing the clothing of the model is not available. In this case, we train our image synthesizer module (ISM) using a self-supervised training strategy. An example of training data of this module is illustrated in Fig.~\ref{fig: training_data_ism}.
The objective of our self-supervised training is to make the network of ISM learn to fit the model's segmented clothes back onto him. Therefore, for a given model image we first remove all the upper body details from it. However, during testing, if the model and the person are wearing clothes of different structural shapes (e.g., jacket and vest), the network may generate artifacts in the regions of the previous clothes of the person which do not overlap with the clothing regions of the target warp. This is because such a region does not exist in the training data. To overcome this, we adopt a couple of tricks. First, while removing the respective clothing details from the model image, we intentionally change the contour of the clothing region of the model image in the training data by dilating the segment. Next, we perturb the segmented model cloth by applying the same perturbation method specified in the training strategy of MGM. 
These are done because due to the self-supervision, the network does not encounter much difficult situations during training. In other words, to make the network robust to warping glitches, we induce such problems artificially during training. We then combine the perturbed clothing segment with the modified model image.
Fig.~\ref{fig: training_data_ism} illustrates the preparation of training data of ISM step by step for better understanding.
Note that the mask of the clothing segment is also provided as input. This makes the network understand the true clothing area of the target. Along with this, the densepose representation of the model is also provided that aids the network in understanding the body shape under clothes.

It should be noted that all three modules: PGWM, MGM, and ISM are trained independently of each other. This has two advantages. First, unlike the previous methods~\cite{cpvton, multiposevton, vtnfp, viton} our modules can be trained in parallel. This is advantageous in terms of training time in case sufficient resource is available. Second, any training error of one module does not affect the training of other modules. In other words, each module is trained optimally.
\section{Experimental Results}
\label{experiments} 

In this section, we first introduce the experimental details of the proposed method, and then we present a comparative study of LGVTON (our method) with other comparing methods such as CP-VTON~\cite{cpvton}, VITON~\cite{viton}, VTNFP~\cite{vtnfp}, MG-VTON~\cite{multiposevton} and M2E-TON~\cite{m2e} both qualitatively and quantitatively. 
Due to the unavailability of the official implementation of VTNFP, it is compared quantitatively based on our implementation following parameter settings suggested in the paper. 

\subsection{Dataset} 
\label{dataset}
We have reported our results on two datasets: the In-shop clothes retrieval dataset of DeepFashion~\cite{deepfashion} and the MPV dataset~\cite{multiposevton}. In-Shop Clothes Retrieval benchmark dataset contains multiple views of each person (front, side, back, and full). It has in total 52,712 images and each image is annotated with fashion landmarks for either upper body or lower body. Since we are focusing on upper body clothes only, so we did the experiments on 33,536 upper body annotated samples. We have randomly selected 3000 pairs of images (with front pose only, to maintain consistent comparison with all comparing methods) from this dataset for testing purposes. Note that the performance of our method on side and back poses is also discussed later.

MVP dataset~\cite{multiposevton} contains in total 35,687 images. For each image, the corresponding human segmentation, human pose estimates, and the corresponding separate clothes images are also provided. Since the baseline methods~\cite{viton, cpvton, vtnfp, multiposevton} require separate clothing images which are not available in DeepFashion, we have done a comparative study on this dataset only. For the comparative study, a set of 100 images are selected from this dataset and annotated manually with fashion landmarks. From these images, we randomly selected 3000 image pairs for testing. Some sample images of this dataset can be viewed in the 1st and 2nd columns of Fig.~\ref{fig: results_compare2}. 

\subsection{Quantitative Comparison}
\label{quantitative} 

For quantitative evaluation including comparison with other methods, we have reported the scores obtained with two metrics, namely, Fréchet Inception Distance (FID)~\cite{fid} and SSIM~\cite{ssim}. 
Note that there is no specific metric related to virtual try-on. However, recent methods~\cite{multiposevton, viton} have used Inception score (IS)~\cite{is} and SSIM~\cite{ssim}. While both Inception score (IS) and Fréchet Inception Distance (FID)~\cite{fid} are evaluation metrics of GAN but unlike FID, IS does not consider real data at all. As a result, IS can not estimate how well the generator approximates the real data distribution. 

Fréchet Inception Distance (FID) is a measure of similarity between two sets of images. It extracts the features embedded in both real and generated images from a layer of inception v3 model~\cite{inceptionv3} pretrained on ImageNet~\cite{imagenet}. Considering the embedding as a continuous multivariate Gaussian, the mean and covariance are estimated for both the generated ($\mu_{g}$, $\sigma_{g}$) and the real data ($\mu_{r}$, $\sigma_{r}$). Then the FID is computed as: $\left\lVert\mu_{r} − \mu_{g}\right\rVert_{2}^{2} + Tr( \sigma_{r} + \sigma_{g} − 2 ( \sigma_{r}\sigma_{g} )^{1/2} )$. A lower value of FID indicates better results.
\begin{table}[h]
	\centering
	\caption{Quantitative evaluation on different datasets.}
	\begin{tabular}{llcc}
		\toprule
		Dataset&Methods & FID$\downarrow$ & SSIM$\uparrow$\\
		\midrule
		DeepFashion & LGVTON (ours) & 56.23 & 0.86\\
		\midrule
		\multirow{5}{*}{MPV} & VITON & 82.43 & 0.74\\
		&CP-VTON & 77.45 & 0.79\\
		&VTNFP & 74.62 & 0.80\\
		&MG-VTON & 70.52 & 0.71\\
		&LGVTON (ours) & \textbf{56.11} & \textbf{0.89}\\
		\bottomrule
	\end{tabular}
	\label{table_is_fid}
\end{table}

The values of FID due to different methods are shown in Table~\ref{table_is_fid}. We also report SSIM scores in the same table. The SSIM scores are for the images of the training set only, since ground-truth is not available for the test set of images. The results demonstrate that our method outperforms others in terms of both metrics.
\subsection{Qualitative Comparison}
\label{qualitative} 
\subsubsection{Visual Comparison} 
\label{visual_comparison}
Here, we present a visual comparative study of our method (LGVTON) with some baseline algorithms, such as VITON~\cite{viton}, CP-VTON~\cite{cpvton}, M2E-TON~\cite{m2e} and MG-VTON~\cite{multiposevton} in Figs.~\ref{fig: results_compare} and~\ref{fig: results_compare2}~\footnote{More results of LGVTON are given in the Appendix.}.
\begin{figure}[!h]
	\centering
	\includegraphics[scale=0.58]{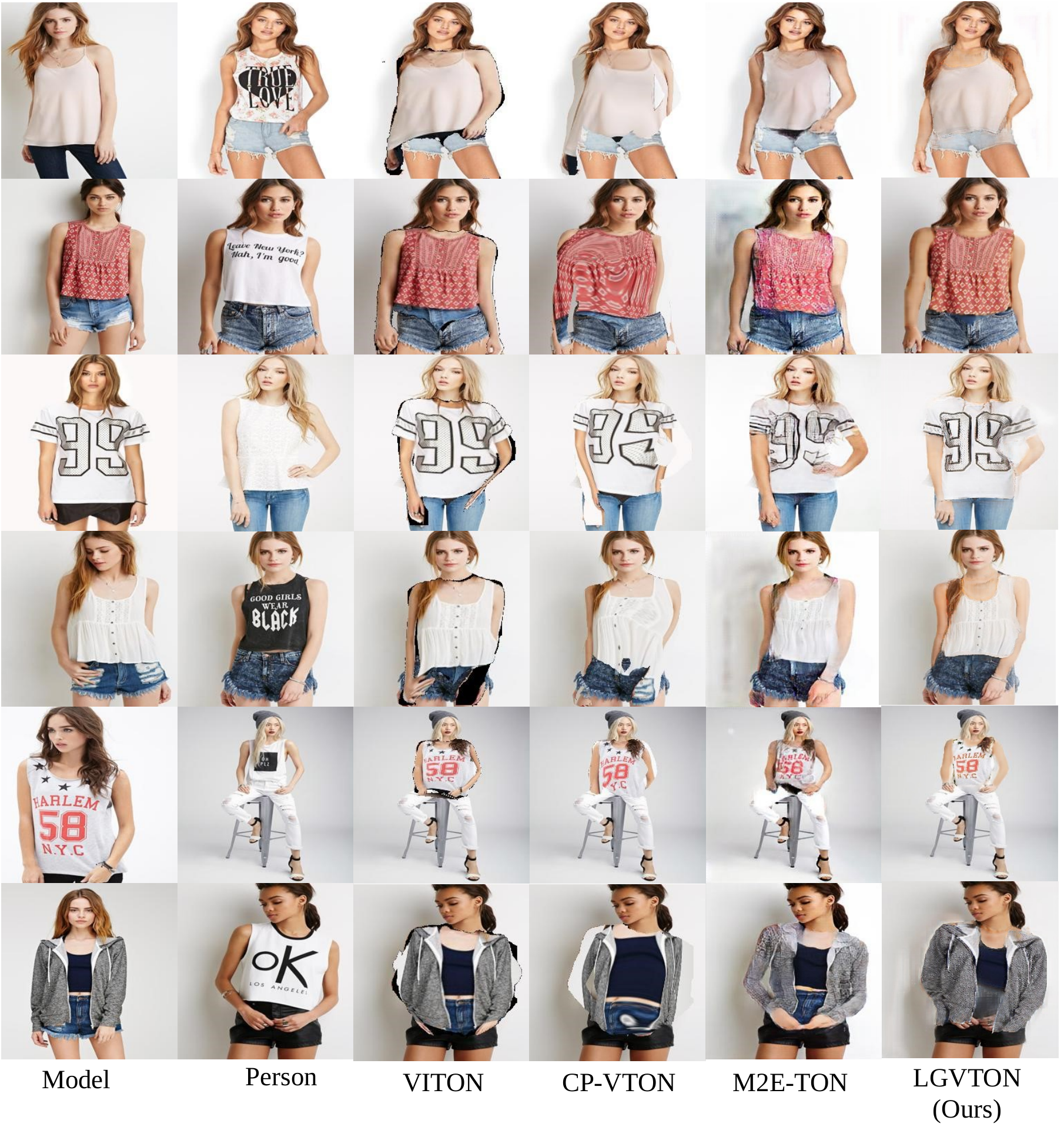}
	\caption{Qualitative comparison with other methods. First five columns are taken from M2E-TON paper.}
	\label{fig: results_compare}
\end{figure}
\begin{figure}[h]
	\centering
	\includegraphics[scale=0.62]{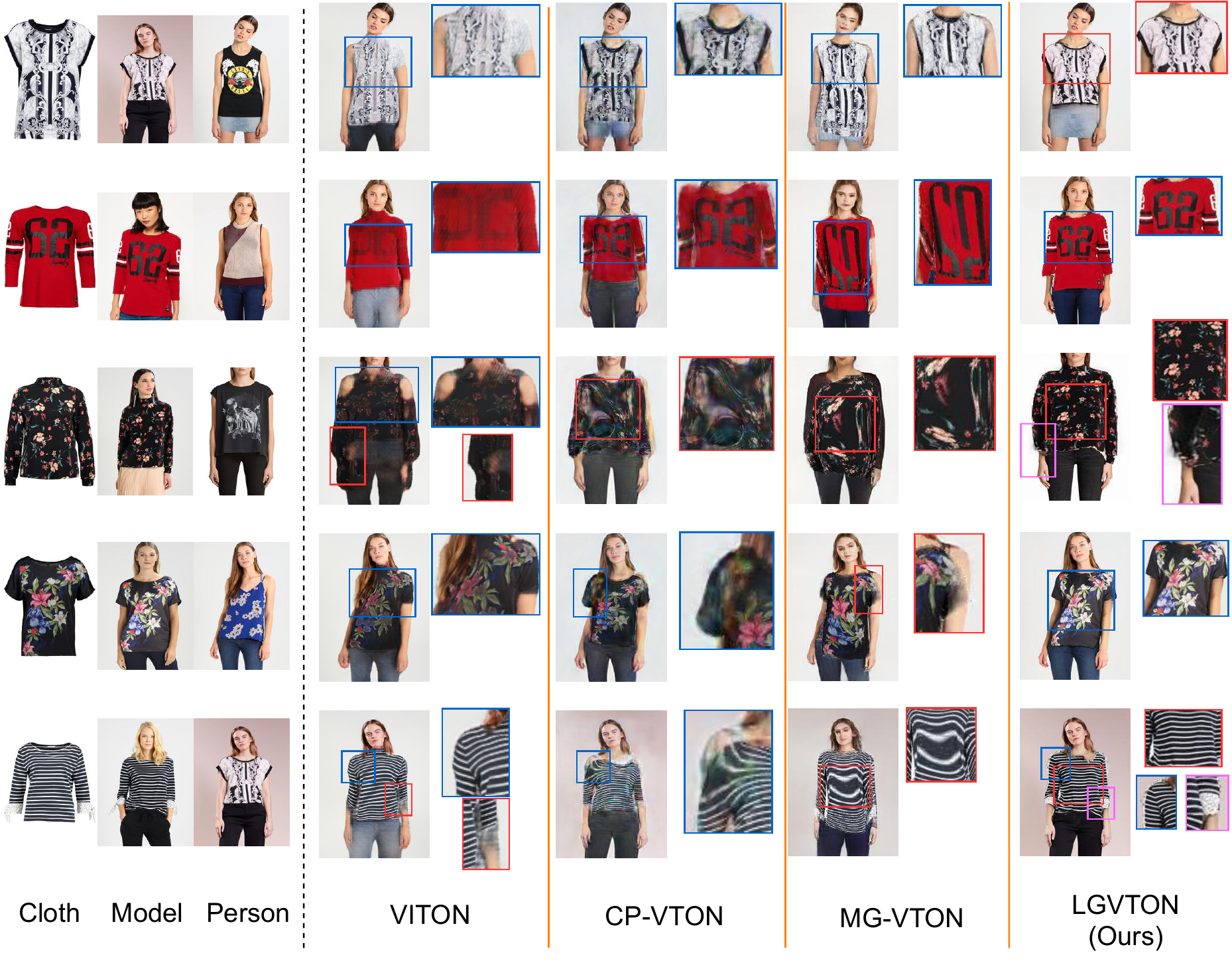}
	\caption{Comparative study on MPV dataset. Significant details are zoomed-in and shown right after each output.}
	\label{fig: results_compare2}
\end{figure}

\begin{figure}[h]
	\centering
	\includegraphics[scale=1.3]{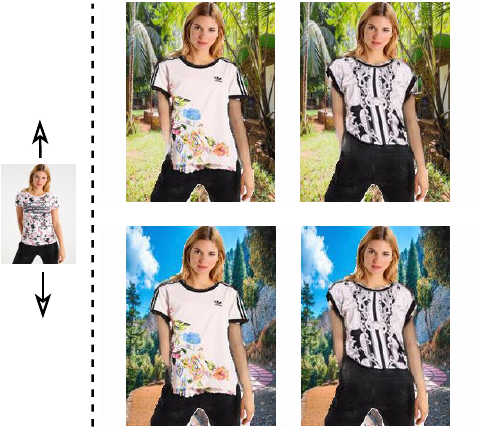}
	\caption{Results on images in cluttered background. Observe that the performance of LGVTON is unaffected by the background clutter.}
	\label{fig: results_clutter_bkg}
\end{figure}
It is observed that in terms of preserving the person and clothes details, LGVTON performs better compared to the baselines. In the results of M2E-TON (Fig.~\ref{fig: results_compare}), it can be observed that the colors of the target clothes are not the same as that of the model clothes and the faces look brighter. However, no such artifact or photometric change is observed in our results. In both Fig.~\ref{fig: results_compare} and Fig.~\ref{fig: results_compare2} it is observed that based on the quality of the target warp, the results of VITON come after LGVTON, which indicates that the point correspondence-based methods (both LGVTON and VITON) perform better compared to geometric matching (e.g., CP-VTON and MG-VTON). 
The geometric matching network employed in both CP-VTON and MG-VTON learns the TPS transformation of the clothes to its corresponding target warp. This network learns the transformation of the mesh grid of the source clothes considering the grid points as control points. Now, a grid can be deformed in numerous ways. However, the human body undergoes restricted deformations due to the presence of limbs, bone joints, and muscles. Considering this, instead of leaving the entire job of learning the possible set of TPS transformations to a neural network, we choose to restrict the grid transformation implicitly to a feasible range of transformations. To achieve this, our method infers the target control points from a set of source control points, and these points are anatomically meaningful in terms of the human body (human landmarks) as well as clothes (fashion landmarks). In other words, the constraint on the possible set of transformations is implicitly imposed in our method.

The problem of loss of details of other clothes, e.g., pants, etc., is evident in the results of CP-VTON and VITON. On the other hand, the results of MG-VTON show a poor reconstruction of the original person details. For example, face details are not well retained by MG-VTON. Based on the overall quality of the outputs, we observe that LGVTON outperforms the other methods, which is due to the combined effect of (i)~landmark guided warping of the model clothing leading to better target warps and (ii)~the convex combination layer in the ISM that helps to retain the details of the warped clothing and the person in the final try-on output.

We also did some experiments (as shown in Fig~\ref{fig: results_clutter_bkg}) to show that LGVTON's performance is unaffected in the presence of background clutter, which establishes its applicability in an in-the-wild setting. To generate these results, we took a random image from the MPV dataset and modified it by adding backgrounds collected from the internet then applied virtual try-on on it.

\begin{figure}[h]
	\centering
	\includegraphics[scale=0.4]{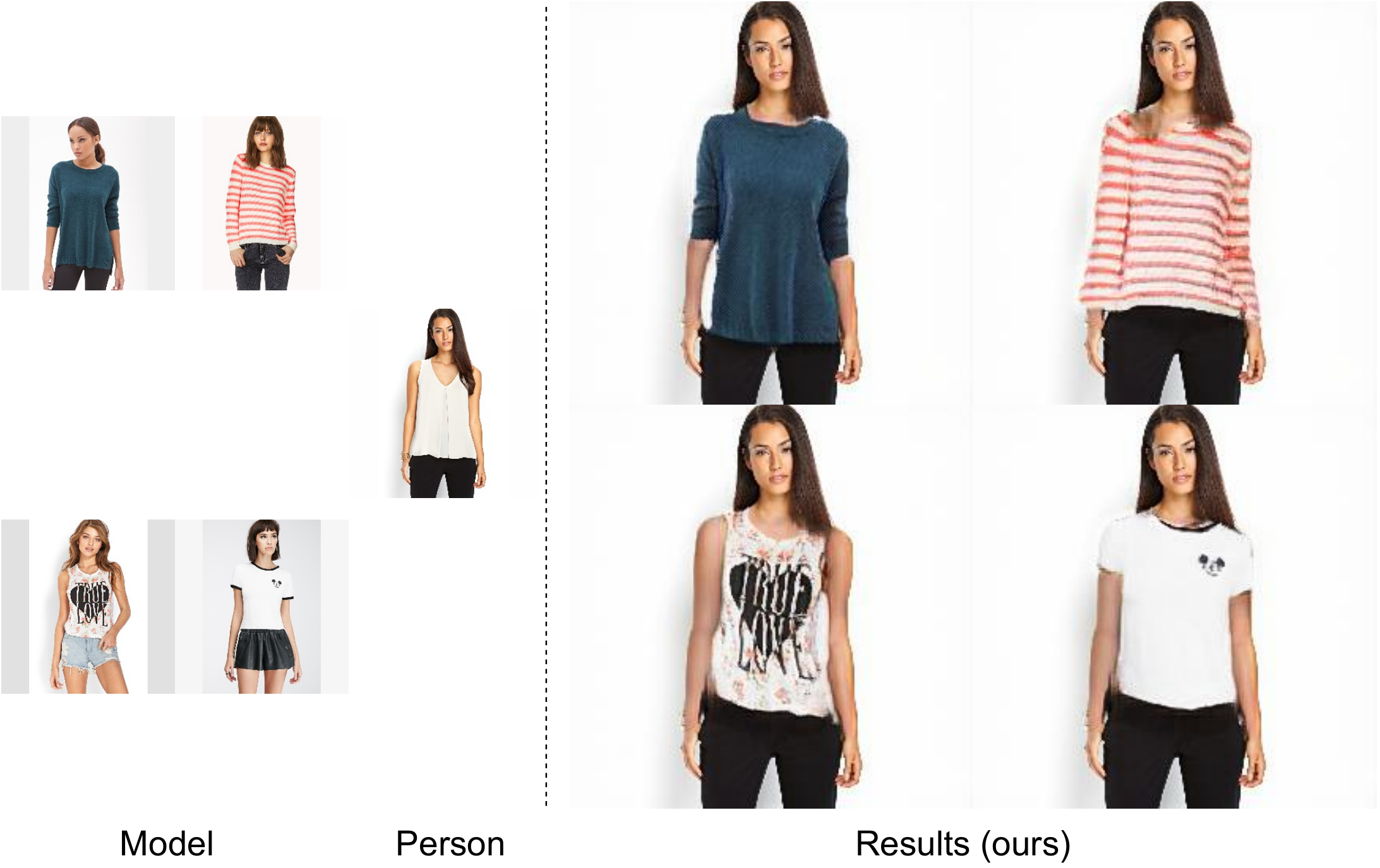}
	\caption{Results of LGVTON on DeepFashion dataset while the training has been done on MPV dataset. This shows that LGVTON works across datasets.}
	\label{fig: results_across_dataset}
\end{figure}

To show that LGVTON works across datasets we tested an instance of LGVTON on DeepFashion dataset while the training is done on the MPV dataset. Some results of this experiment are shown in Fig.~\ref{fig: results_across_dataset}. Note that since MPV does not contain fashion landmark annotations, hence, in this experiment we used the PGWM trained on the DeepFashion dataset. Although we annotated the fashion landmarks for 100 images in MPV for our qualitative and quantitative testing purposes, this is not sufficient for training the fashion landmark predictor network.

\subsubsection{Human Study}
\label{human_study} 

We made a group of 30 volunteers and each person is presented a randomly selected set of $100$ results from the results of all concerning methods on our test set of MPV dataset. Pairwise comparison of the results between each of the comparing methods and our method is done. The result of this study is reported in Table~\ref{table_us}.
\begin{table}[h]
	\centering
	\caption{Human study. Each cell presents the percentage by which LGVTON (our method) is preferred over the comparing method on an average.}
	\begin{tabular}{ll}
		\toprule
		Methods & Human Study$\uparrow$\\
		\midrule
		VITON / LGVTON & 76.23\%  \\
		CP-VTON / LGVTON & 74.72\% \\
		VTNFP / LGVTON & 71.54\% \\
		MG-VTON / LGVTON & 71.21\% \\
		\bottomrule
	\end{tabular}
	\label{table_us}
\end{table} 

\subsection{On Different poses of the Model and the Person} 
\label{experiments_multipose} 

\begin{figure}[h]
	\centering
	\includegraphics[scale=1.3]{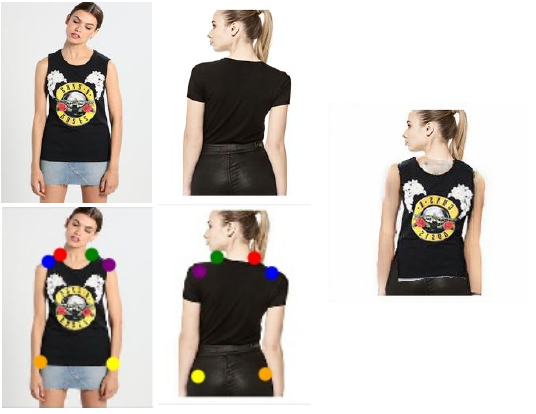}
	\caption{Results on the back pose shows LGVTON is not constrained to front pose only.}
	\label{fig: results_back_pose}
\end{figure}

Here we examine the performance of LGVTON on different poses (other than the front as it has been already discussed) of the model and the person. As shown in Fig.~\ref{fig: results_back_pose} it may be observed that LGVTON is not only constrained to the front pose but works even for the back pose also. 

\begin{figure}[h]
	\centering
	\includegraphics[scale=0.32]{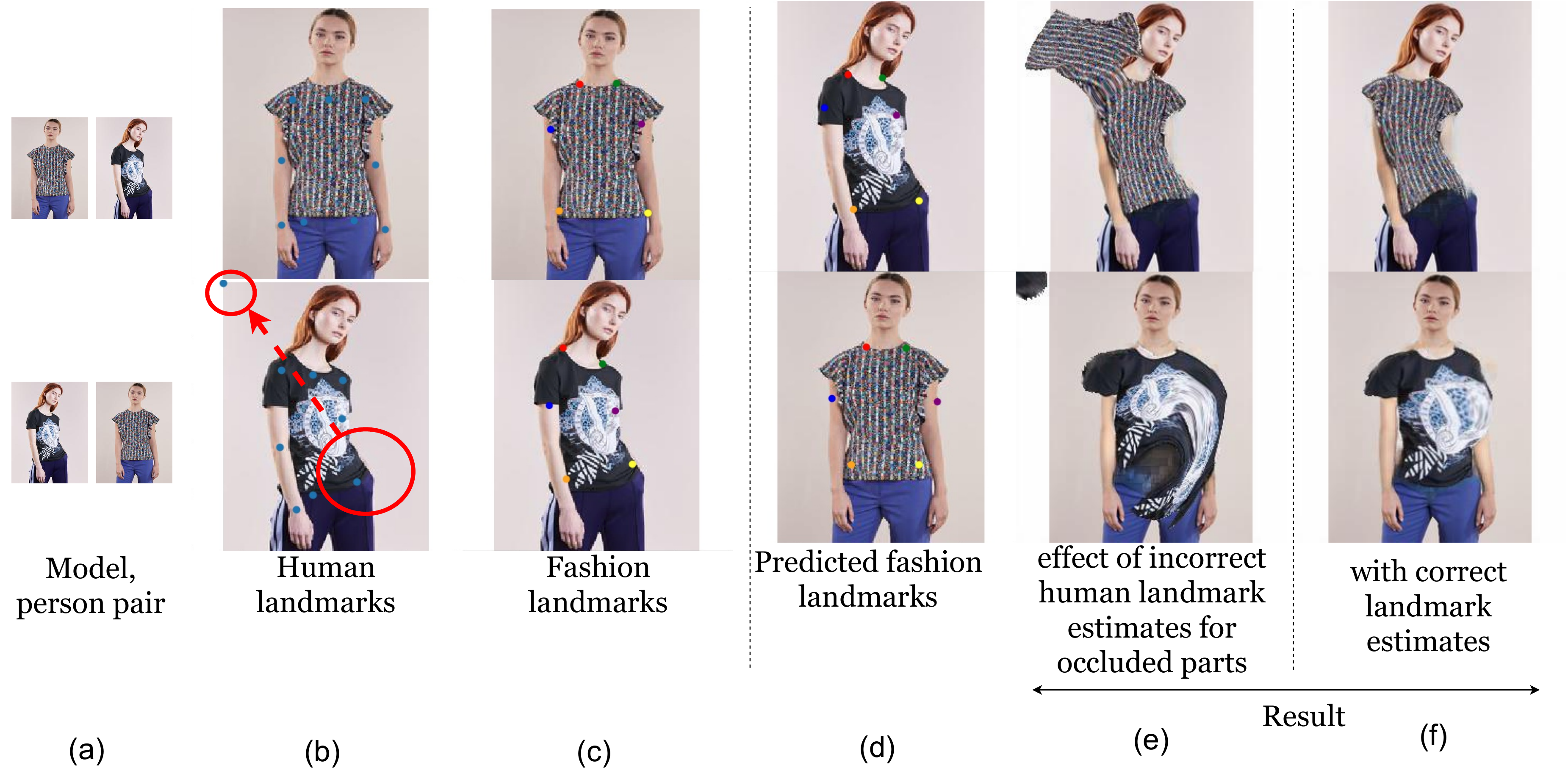}
	\caption{LGVTON shows poor performance in case at least one of the model or the person is in side pose.}
	\label{fig: multi_perspective}
\end{figure}

However, this method is not scalable to the cases when the image of the model or the person or both are in the side view. In Fig.~\ref{fig: multi_perspective} we have demonstrated two cases. In the first case (first row) - the model is in front and the person in the side pose. Here we see that due to the occlusion of the hand of the person, the human landmark there is estimated wrongly (b), which affects the results (e). Although with a correct estimate the result improves (f) but the left sleeve of the target warp that should have been occluded in the result, is visible. In the second case (second row) - the model is in the side pose and the person is in the front pose. Similar to the previous case, an incorrect landmark estimation affects the result (e). However, even with the correct estimates, we see that, instead of interpolating the missing part of the cloth, LGVTON incorrectly stretches the torso part of the clothes in the location of the sleeves (f). These observations suggest that even with correct estimates of both sets of landmarks LGVTON performs poorly in the case of side pose images.

\subsection{Ablation Study}
\label{ablation} 

In this subsection, we present a quantitative study on the significance of each component of LGVTON. This study is conducted on the MPV test set (explained in Section~\ref{dataset}) based on the FID metric. The values of this metric are reported in Table.~\ref{table_ablation}, which shows complete LGVTON achieves the best FID score. However, since it has been observed in existing works~\cite{multiposevton, swapnet, viton} that GAN based evaluation metrics are not very reliable in terms of accurately measuring the quality of VTON output, we have also given a human study in Table.~\ref{table_human_study_ablation} (conducted similarly as mentioned in the Section~\ref{human_study}). This study also indicates that the results of LGVTON are chosen most of the time over the comparing methods~\footnote{\label{supp_ablation}Please see the Appendix for a detailed study on the utility of each component of LGVTON.}.

\begin{table}[h]
	\centering
	\caption{Quantitative ablation on MPV dataset.}
	\begin{tabular}{llcc}
		\toprule
		Methods & FID$\downarrow$\\
		\midrule
		LGVTON (w/o MGM) & 57.27\\
		LGVTON (w/o correlation layer in PGWM) & 57.65\\
		LGVTON (w/o fashion landmarks) & 56.32\\
		LGVTON (w/o fashion landmarks, w/o MGM) & 57.95\\
		LGVTON (ours) & \textbf{56.11}\\
		\bottomrule
	\end{tabular}
	\label{table_ablation}
\end{table}

\begin{table}[h]
	\centering
	\caption{Human Study. Each cell presents the percentage by which LGVTON (our method) is preferred over the comparing method on average.}
	\begin{tabular}{llcc}
		\toprule
		Methods & Human Study$\uparrow$\\
		\midrule
		LGVTON (w/o MGM) & 85.60\%\\
		LGVTON (w/o correlation layer in PGWM) & 91.33\%\\
		LGVTON (w/o fashion landmarks) & 80.10\%\\
		LGVTON (w/o fashion landmarks, w/o MGM) & 87.30\%\\
		\bottomrule
	\end{tabular}
	\label{table_human_study_ablation}
\end{table}

\subsection{Implementation Details}
\label{implementation_details}

We trained PGWM, MGM and ISM approximately for 316, 9000, 20 number of epochs respectively, with Adam optimizer keeping $lr$ = 0.001, beta1 = 0.9, beta2 = 0.999. The network architectures of our modules have been discussed in Appendix. To report the computation cost of our method, below we discuss the running times of each of the components of LGVTON. We also compare the number of parameters of our method as well as that of the different comparing methods.

\textbf{Runtime:} The average prediction time over a randomly selected set of 100 samples from our test set on MPV dataset is reported in Table.~\ref{tab: run_time}. The results are obtained on NVIDIA Titan XP. Note that the implementations of our modules and the pose estimation method are done in the deep learning framework Keras with TensorFlow backend. For the human parsing and densepose estimation, we have used their official implementations which are in the Caffe framework. For the pose estimation method also we have used its implementation available online. We do not compare the execution time of our method with other methods as inference time is highly dependent on the deep learning framework. The implementations of MGVTON~\cite{multiposevton}, CP-VTON~\cite{cpvton}, VITON~\cite{viton} are in the PyTorch framework, while ours is in the Keras framework. Due to differences in frameworks, comparing the execution times would not be justified. It is noteworthy that the implementation times of our modules can be reduced in case applied in time-efficient deep learning frameworks, e.g., Caffe, PyTorch, or TensorFlow.
\begin{table}[h]
	\centering
	\caption{Prediction time of each of the components of LGVTON. Since LGVTON uses different annotations, hence we also report the running times for each of the annotation methods. }
	\begin{tabular}{llcc}
		\toprule
		Methods & Prediction time$\downarrow$\\
		\midrule
		PGWM & 1.32 $\pm$ 0.12ms\\
		MGM & 504.12 $\pm$ 20.82ms\\
		ISM & 599.14 $\pm$ 18.17ms\\
		Human parsing & 120.21 $\pm$ 3.64ms \\
		Pose estimation & 1309.51 $\pm$ 113.08ms\\
		Densepose estimation & 88.42 $\pm$ 3.41ms\\
		\bottomrule
	\end{tabular}
	\label{tab: run_time}
\end{table}

\textbf{Number of parameters:} Number of parameters is an implementation-independent metric that is a reasonable factor for comparing the computation cost. Therefore, we report the number of parameters of our method as well as that of the comparing methods in Table.~\ref{param_comp}. Each of these methods uses human parsing~\cite{lipssl} and pose estimation~\cite{openpose} methods to get the corresponding annotations, whereas our method uses an additional annotation of densepose~\cite{densepose}. We report the count of parameters of~\cite{lipssl, openpose, densepose} in the third column. The total number of parameters of the annotation methods corresponding to each of the compared methods is reported in the fourth sub-column of the third column. Let us call this parameter count type-2. While the number of parameters in the try-on methods excluding that of the annotation method's is reported in the second column. Let us call this parameter count type-1. Finally, the total number of parameters is reported in the fourth column (type-1 + type-2). Note that although the total count of parameters in our method is a little higher than that of VITON and CP-VTON, the type-1 parameter count is quite less in our method compared to others. The number of parameters in our warping module is 1.69M compared to that of CP-VITON's 19.05M and MGVTON's 45.80M. We do not compare VITON here as it uses shape context matching based TPS warping~\cite{shape_context}, a traditional warping method that has its limitations. However, the point we want to make here is that our way of warping reduces the computation cost of warping significantly.
\begin{table}[!h]
	\caption{Comparing the number of parameters of different methods.}
	\begin{center}
		\begin{tabular}{ccccccc}
			\toprule
			\multirow{2}{*}{Methods} &
			\multirow{2}{*}{\#Parameter (type-1)} &
			\multicolumn{4}{c}{\#Parameter} &
			\multirow{2}{*}{Total} \\
			&
			&
			\begin{tabular}[c]{@{}c@{}}Human \\ parsing\end{tabular} &
			\begin{tabular}[c]{@{}c@{}}Pose \\ estimation\end{tabular} &
			\begin{tabular}[c]{@{}c@{}}Densepose \\ estimation\end{tabular} &
			\begin{tabular}[c]{@{}c@{}}Total \\ (type-2)\end{tabular}
			\\
			\midrule
			VITON   & 29.34M  & 75.65M & 52.31M & -      & 127.96M & 157.30M \\
			CP-VTON & 40.40M  & 75.65M & 52.31M & -      & 127.96M & 168.36M \\
			MGVTON  & 224.46M & 75.65M & 52.31M & -      & 127.96M & 352.42M \\
			LGVTON  & 2.45M   & 75.65M & 52.31M & 59.73M & 187.69M & 190.14M\\
			\bottomrule
		\end{tabular}
		\label{param_comp}
	\end{center}
\end{table}
\section{Conclusion}
\label{conclusion} 
This paper presents a self-supervised landmark guided approach to virtual try-on which synthesizes the image of a person wearing model clothes. Unlike many existing works, this work requires only the images of the model and the person without requiring any separate clothes image, which makes it more effective, as having a separate clothes image is difficult. Our method contains three modules. The first module utilizes the correspondence between the estimated landmark sets of the model and the person to predict the target warping of the model clothes. We propose a mask generator module that is responsible for refining the fit of the warped clothes in the virtual try-on output. Our final module, the image synthesizer module, combines the aligned model clothes and person to synthesize the final output. We conducted an ablation study that establishes the necessity and efficacy of each of these modules. Comparison with four existing methods on two datasets using two performance measures demonstrates the ability of this method in generating perceptually convincing virtual try-on outputs. In most of the experiments, our methods outperform the state-of-the-art algorithms. 
This is worth mentioning that our work explores a new research direction involving landmarks in the domain of virtual try-on. 

\begin{figure}[h]
    \center
	\includegraphics[scale=0.63]{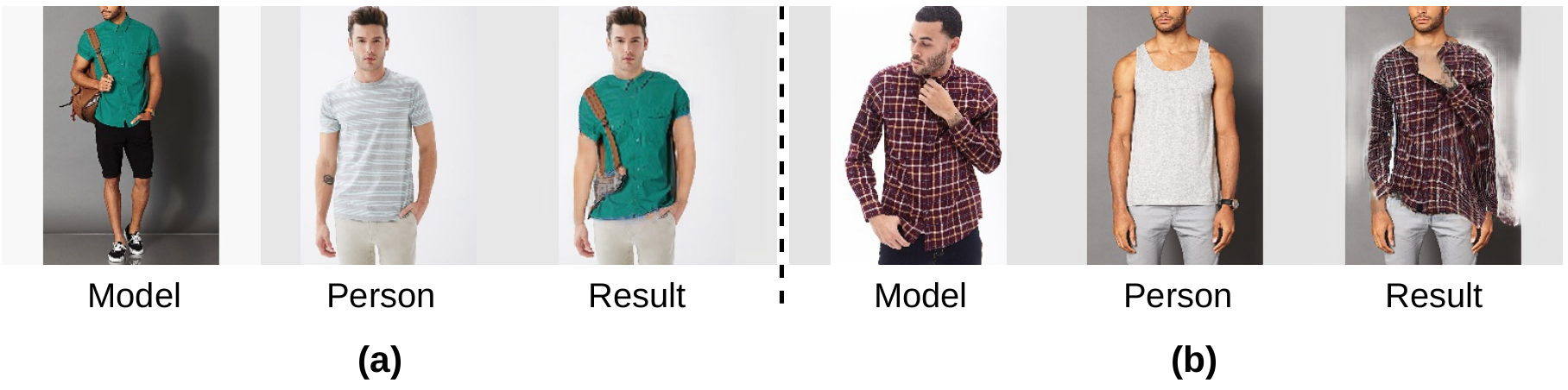}
\caption{Failure cases of LGVTON. (a)~ Incorrectly segmented model clothes (model's hand and bag strap are segmented as clothes incorrectly) result in poor VTON output. (b)~Issues with folded clothe sleeves. As seen here, the left sleeve of the model clothes is cross folded. Which could not be unfolded in the result to warp according to person pose.}
\label{fig: failure}
\end{figure}

Our method has a few limitations too. First, LGVTON is sensitive to the quality of segmentation~\cite{lipssl} of the model image (see Fig.~\ref{fig: failure}(a)).
Second, LGVTON can not handle the folding of clothes with long sleeves (see Fig.~\ref{fig: failure}(b)). This indicates the proposed method is restricted to less complex hand postures. However, this is due to the thin-plate spline based warping method on the 2D mesh grid. It is also an issue with the existing virtual try-on methods using similar transform and a big challenge in this domain of work. Third, this method is restricted to the front and back view of human images only. In the future, we plan to investigate more on these issues. In addition, we believe, some aspects of this work are worth exploring more. First, the issue of warping glitches. We have already discussed that the human landmarks are not sufficient representations of body shape, but, pose, which results in warping glitches. However, we infer that the landmarks representing both body shape and pose together might solve the issue of warping glitch leading towards a better solution. Second, we have worked with only 6 fashion landmarks in this work. However, we argue that a dense set of fashion landmarks encoding the fine-grained clothing details might produce better results. We believe deeper research towards these possible directions might lead to improved solutions.

\section*{Declarations}
\subsection*{Funding}
Not applicable.
\subsection*{Conflict of interest}
The authors declare that they have no conflict of interest.
\subsection*{Availability of data and material}
The datasets used in this work are available in public domains. The sources are appropriately referred to. 
\begin{acknowledgements}
The authors sincerely thank Aruparna Maity and Dakshya Mishra for their help and thank Sankha Subhra Mullick for helpful discussion during this work.
\end{acknowledgements}
\appendix
\section{Appendix}
In this appendix, first, we discuss Thin-plate spline (TPS) transformation. Second, we give a detailed study on the idea of the correlation layer that has been used in the PGWM of LGVTON. Third, we give a detailed ablation study to clarify in detail the effectiveness of each of the components of LGVTON. Fourth, we provide the implementation details of our network architecture. In addition, we give some results of LGVTON on DeepFashion~\cite{deepfashion} and MPV dataset~\cite{multiposevton} in Figs.~\ref{fig: more_results_df1},~\ref{fig: more_results_mpv1},~\ref{fig: more_results_mpv2}, and~\ref{fig: more_results_mpv3}.

\subsection{Thin Plate Spline (TPS)}
Given a set of pair of source and target landmarks \{($\mathbf{r}_j$, $\mathbf{t}_j$); $j$ = 1, . . . , $N$\}, a polyharmonic smoothing spline fits a function $f(\cdot)$ between the sets \{$\mathbf{r}_j$; $j$ = 1, . . . , $N$\} and \{$\mathbf{t}_j$; $j$ = 1, . . . , $N$\} that minimizes the following objective functional \cite{polyharmonic}:
\begin{equation}
	\tau[f] = \sum_{j=1}^N \|f(\mathbf{r}_j)-\mathbf{t}_j\|_2^2 + \iint\limits_{\mathbb{R}^2} \|\nabla^m f\|_2^2dx\ dy 
	\label{tps1} 
\end{equation}
where $\nabla^m f$ is the vector of all $m$-th order partial derivatives of $f$ that enforces smoothness on $f$. In our case, $\mathbf{r}_j, \mathbf{t}_j\in \mathbb{N}^{2}$, or in other words, $\mathbf{r}\equiv(r^x, r^y)$ and $\mathbf{t}\equiv(t^x, t^y)$. 
A special case of polyharmonic smoothing spline where $m$ = 2, called thin-plate spline (TPS), is proposed by Duchon \cite{duchon}. It minimizes the following objective functional based on eq.~\eqref{tps1}: 
\begin{equation}
	\tau[f] = \sum_{j=1}^N \|f(\mathbf{r}_j)-\mathbf{t}_j\|_2^2 + \iint\limits_{\mathbb{R}^2} [f_{xx}^2 + 2f_{xy}^2 + f_{yy}^2]dx\ dy 
	\label{tps2} 
\end{equation}
The physical analogy to the concept of the thin-plate comes from the minimization of second-order gradients ($f_{xx}$, $f_{xy}$, $f_{yy}$) in its objective, which restricts bending and enforces smoothness in the TPS fit. This is similar to the effect of physical rigidity in the thin metal plate.

A closed form solution of it as proposed in \cite{wabha} is given by  
\begin{equation}
	\mathbf{t} = f(\mathbf{r}) = \mathbf{a}_0 + \mathbf{a}_1 r^x + \mathbf{a}_2 r^y + \sum_{j=1}^N \mathbf{c}_j\phi(\|\mathbf{r} - \mathbf{r}_j\|_2),
	\label{tps3} 
\end{equation}
where $\mathbf{a}_0$, $\mathbf{a}_1$, $\mathbf{a}_2$, $\mathbf{c}_j$ are parameters with dimension equal to the dimension of landmarks, which is 2 in our case.

The TPS is represented in terms of radial basis function (RBF) $\phi$. Given a set of control points {$\mathbf{r}_{j}$, $j=1, ..., N$} a RBF maps a given point $\mathbf{p}$ to a new location $g(\mathbf{p})$, represented by, 
\begin{equation}
	g(\mathbf{p}) = \sum_{j=1}^N \mathbf{c}_j\phi(\|\mathbf{p} - \mathbf{r}_j\|_2)
\end{equation}
The radial basis kernel used in TPS is $\phi(\mathbf{r})$ = ($\mathbf{r}^2$ $ln$ $\mathbf{r}$). 

When the values of $\mathbf{t}_j$ are noisy (due to landmark localization errors), which is a very obvious situation in practice, the interpolation requirement is relaxed by means of regularization. This reduces the problem to an approximation problem~\cite{approximatetps1, approximatetps2}. This is obtained by minimizing
\begin{equation}
	H[f] = \sum_{j=1}^N \|f(\mathbf{r}_j)-\mathbf{t}_j\|_2^2 + \lambda \iint\limits_{\mathbb{R}^2} [f_{xx}^2 + 2f_{xy}^2 + f_{yy}^2]dx\ dy,
\end{equation}
where $\lambda$ is the regularization parameter, which is a +ve scalar.  
Here, $\lambda$ determines the relative weight between the approximation behavior and the smoothness of the transformation. In the limiting case, $\lambda$ = 0, we get an interpolation transformation. If the value of $\lambda$ is small we get a good approximation behavior; for a higher value of $\lambda$, we get a very smooth transformation function. But then the local deformations determined by the set of landmarks are maintained poorly.
%
\subsection{Correlation Layer}
Correlation is a statistical technique that can show whether and how strongly pairs of variables are related. We use the idea of correlation in our Pose Guided Warping Module (PGWM). Here we model the warping of the source clothes as a function of the correlation between the human landmarks of the model and the person, and fashion landmarks of the source clothes. The reason for using the correlation is that when a person wears clothes it gets molded according to his body shape and pose. Now, here we are transferring the clothes from model to person hence, the clothes have to undergo deformation according to the way the body shape and pose changes from model to person. This change is modeled by correlation.

Coming into more detail, given the data $f_a, f_b \in \mathbb{R}^{1\times l}$, the correlation layer produces the correlation map $C_{ab} \in \mathbb{R}^{l \times l}$, where, 
\begin{equation}
	C_{ab} = {f_a}^{T}{f_b}.
\end{equation}
Therefore a correlation map basically contains pairwise similarity of all the values of $f_a$ and $f_b$. This is useful in establishing relationships among different landmarks of model and person to model the warping.
\subsection{Ablation Study}
In this subsection, we conduct an in-depth qualitative study on the significance of each component of LGVTON. 
\subsubsection{Utility of Different Loss Functions for Training ISM}
\begin{figure}[!h]
	\centering
	\includegraphics[scale=0.68]{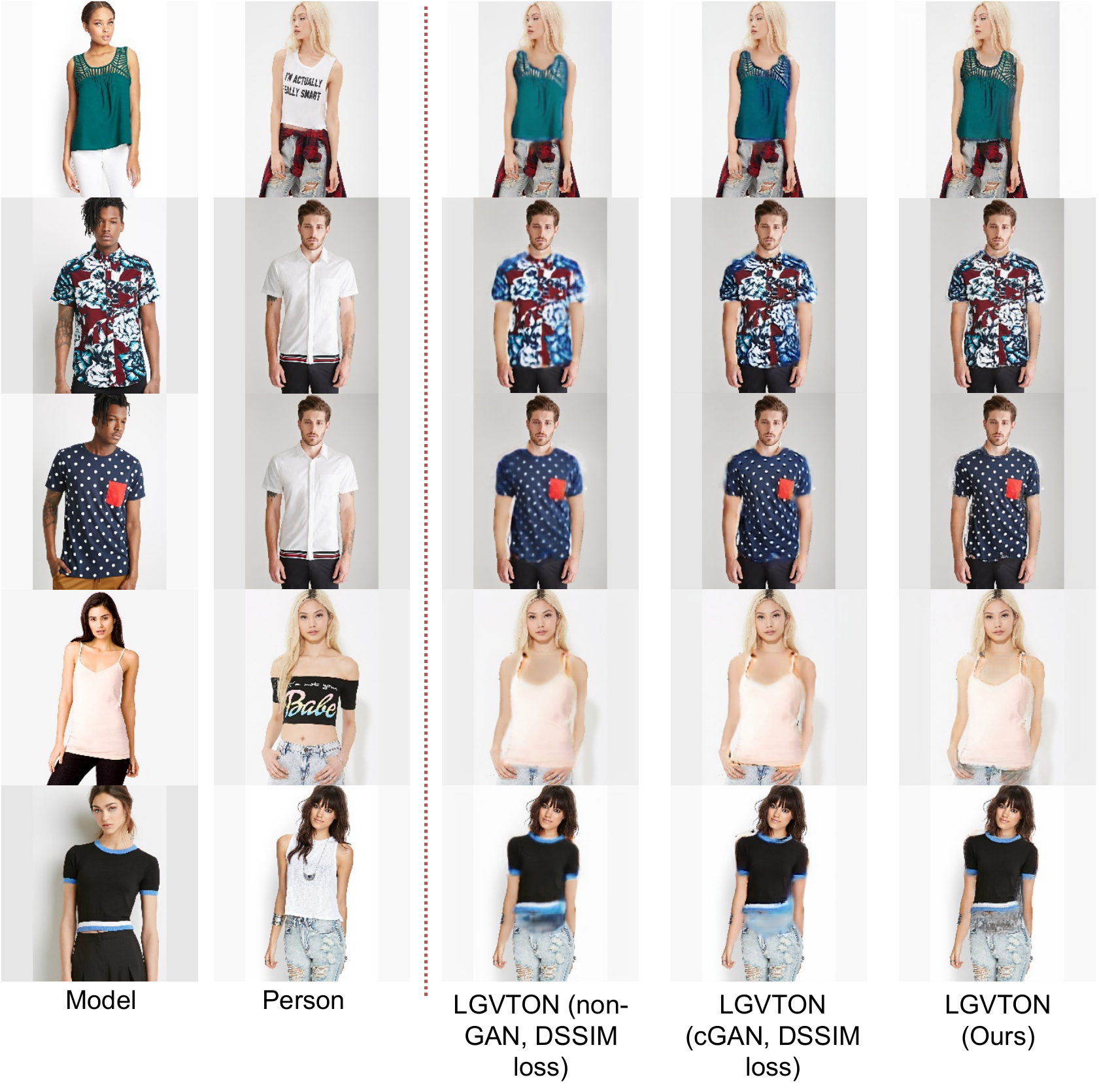}
	\caption{Effectiveness of different losses during training the Image Synthesizer module (ISM). (1)~LGVTON (non-GAN, DSSIM loss) - a non-GAN variant of ISM trained with DSSIM loss, (2)~LGVTON (cGAN, DSSIM loss) - cGAN with the generator trained with DSSIM loss, (3)~LGVTON - similar to (2) but the generator is trained with DSSIM and VGG perceptual loss both. Notice the clarity of output increases from LGVTON (non-GAN, DSSIM loss) to LGVTON.}
	\label{fig: ablation_gan_vs_cnn}
\end{figure}

We run a comparative study on the effect of different loss functions used to train ISM. Keeping the other settings same, we train 3 different instances of ISM with different combinations of loss functions as shown in Fig.~\ref{fig: ablation_gan_vs_cnn}.
As GAN tries to approximate the data distribution, so it generates better output than its non-GAN variant~\cite{gan}. This is evident in the results of the non-GAN variant of LGVTON where the ISM is trained with DSSIM loss. We call this variant LGVTON (non-GAN, DSSIM loss) and the corresponding GAN variant LGVTON (cGAN, DSSIM loss). Comparing the results of LGVTON(cGAN, DSSIM loss) and LGVTON (ours) (a cGAN, where the generator is trained with DSSIM and perceptual loss both), it can be observed that the result improves in the presence of VGG perceptual loss.

\subsubsection{Significance of Fashion Landmarks in PGWM}
\begin{figure}[!h]
	\centering
	\includegraphics[scale=0.63]{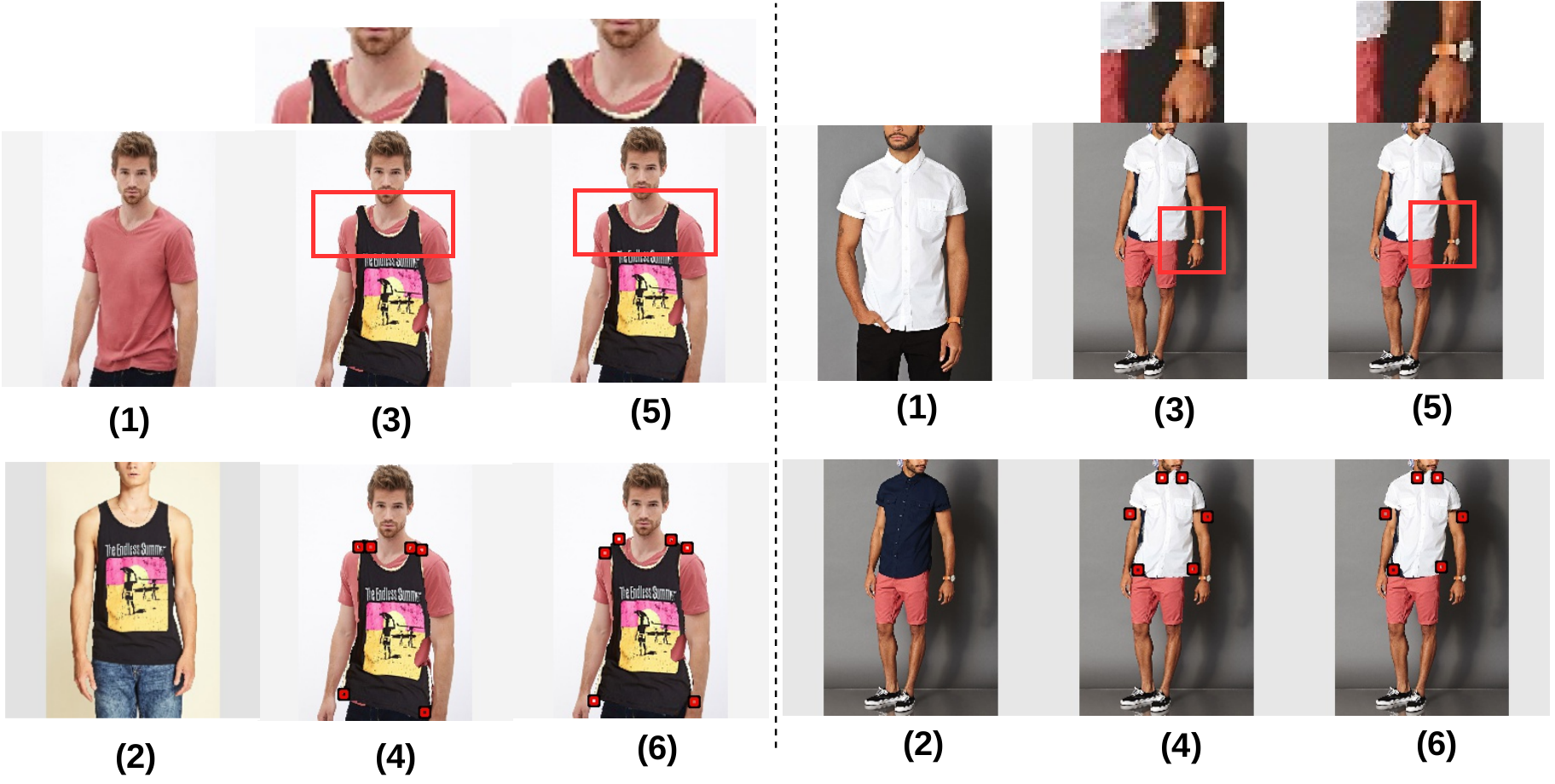}
	\caption{Role of different fashion landmarks in predicting target warp of the model clothes in PGWM. The figure shows the utility of fashion landmarks around the collar (left) and hem (right) respectively. For a better understanding of the viewer, instead of showing the generated warped clothes image only, we show the overlay of it on the person image. (1)~Person, (2)~model, (3)~clothes warped using only human landmarks, (4)~transformed locations of fashion landmarks of the model clothes, obtained by the corresponding transformation function of (3), (5)~clothes warped by PGWM using both human and fashion landmarks, (6)~fashion landmarks predicted in PGWM, which is observed to be more accurate than (4). This results in better warping of the model clothes as shown in (5) in comparison to that in (3).}
	\label{fig: ablation_flm}
\end{figure}

We conduct a study on PGWM when the warping is done with human landmarks only instead of both human and fashion landmarks. Fig.~\ref{fig: ablation_flm} shows two cases portraying the effectiveness of fashion landmarks around the collar and hem. Having only human landmarks might serve the purpose but at the cost of an increase in the amount of warping glitches, which is tackled to some extent in ISM with the help of the mask generated by the MGM. The scores reported in Table.~\ref{table_ablation} and Table.~\ref{table_human_study_ablation} (LGVTON (w/o fashion landmarks)) shows that without fashion landmarks the performance of LGVTON degrades. However, for obvious reasons, the performance degrades even more if the support of the MGM is also removed (observe the score of LGVTON (w/o fashion landmarks, w/o MGM) in Table.~\ref{table_ablation} and Table.~\ref{table_human_study_ablation}).
\subsubsection{Effectiveness of Correlation Layer in PGWM}
\label{ablation: correl}
\begin{figure}[!h]
	\centering
	\includegraphics[scale=0.65]{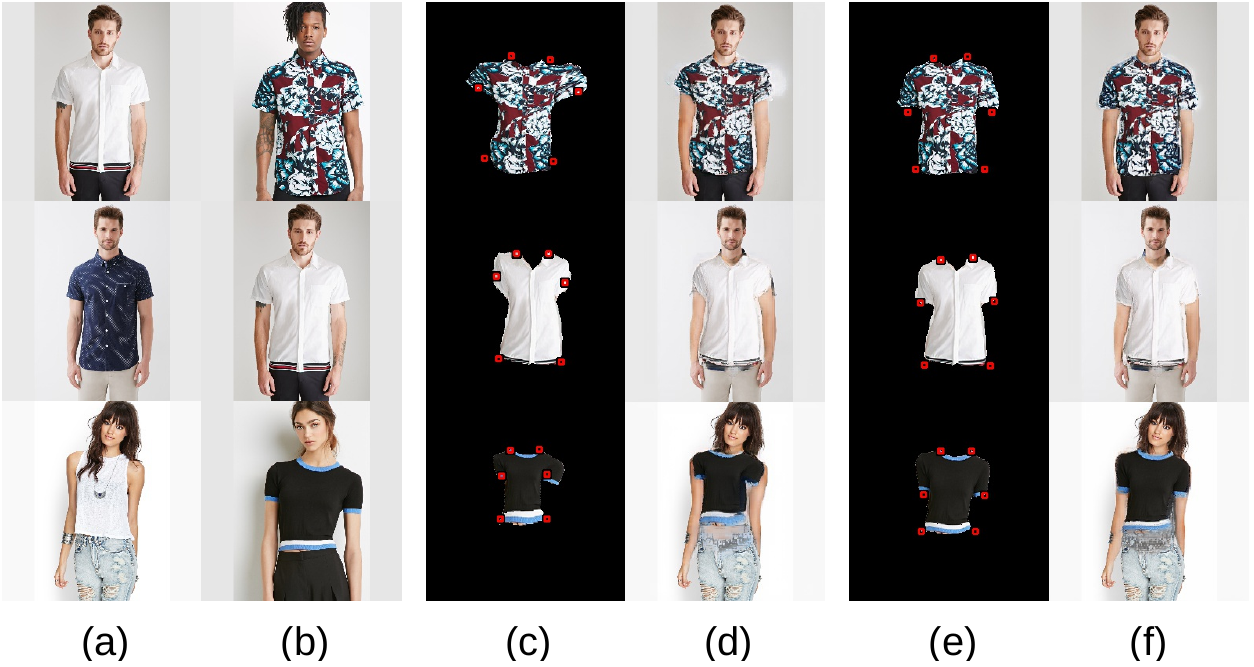}
	\caption{A study on the effectiveness of correlation layer in fashion landmark predictor network $\mathcal{F}$ of PGWP in LGVTON. (a)~Person image, (b)~Model image, (c, e)~predicted locations of fashion landmarks and warp clothes generated by PGWM when $\mathcal{F}$ is trained w/o correlation layer and with correlation layer respectively. (d, f)~final VTON result for the warp cloths shown in (c) and (e) respectively. We observe that the results in (f) are better than that in (d), which justifies the potency of the correlation layer in PGWM.}
	\label{fig: ablation_woCorrel}
\end{figure}

We study the effectiveness of the correlation layer in the fashion landmark predictor network of PGWM (refer to Fig.~\ref{fig: ablation_woCorrel}). The presence of the correlation layer establishes the relationship between the human poses of the model and the person, which in turn assists in predicting better estimates of fashion landmarks of the target warp clothes. 
\subsubsection{Studying the Role of Target Mask as Input to ISM}
\begin{figure}[h]
	\centering
	\includegraphics[scale=0.36]{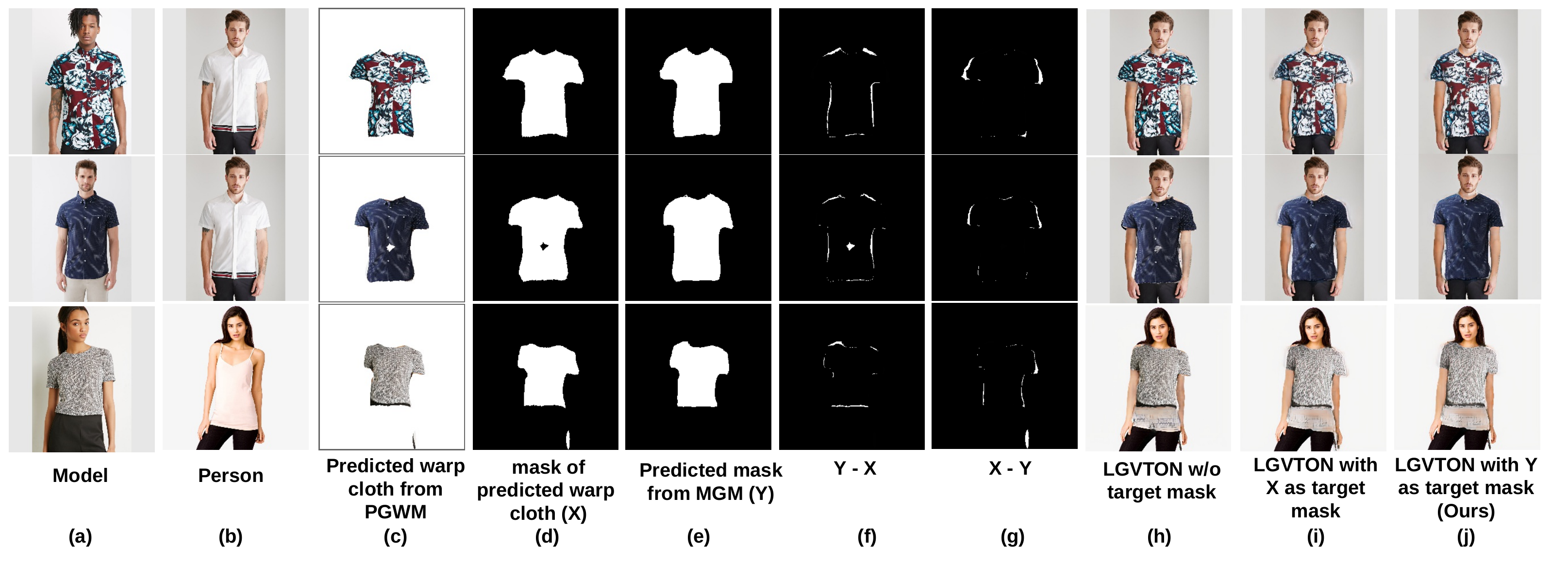}
	\caption{Effectiveness of target mask generated by our mask generator module (MGM). We notice significant differences (as shown in (f), (g)) between the mask (d) of the predicted warped clothes (c) and the mask (e) predicted by MGM corresponding to the warp clothes (c). (h) shows the final try-on result contains artifacts when ISM is trained without the target mask. To show the effectiveness of the target mask we also show the result where instead of (e) we give (d) as target mask input to ISM; which is shown in (i). As we observe the artifacts still remain in (i). Whereas, when (e) is given to ISM as target mask, the result (j) improves, e.g., the areas with pixel value 1 in (f) are filled with necessary color and texture details, and those in (g) are replaced with background details resulting in a better VTON result. Note that the hole in the warp clothes in row 2 is not due to a warping glitch. This is due to inaccurate human parsing. However, LGVTON handles this also.}
	\label{fig: ablation_target_mask}
\end{figure}

For understanding the effect of the target mask, we train an instance of ISM without providing the target mask as input. Now it is observed that w/o target mask, the network can not identify the warping glitches. This can be verified from Fig.~\ref{fig: ablation_target_mask}(h) where the effect of the warping glitch is propagated to the output. However, when a target mask is given as input, ISM can identify the areas of warping glitch and takes the necessary action according to the type of the glitch. The reason being the variety of clothing types makes the network confused to distinguish a warping glitch from the design of the clothes, e.g., in the first two rows, the inappropriate estimation of $c'_{flm}$ causes the sleeves to be stretched more outwards. While that in Fig.~\ref{fig: ablation_target_mask}(j) is not observed as the network removes those areas of extra stretch and replaces them with the background. In the example of the third row Fig.~\ref{fig: ablation_target_mask}(h), the effect of warping glitch exposes some body parts near the right neckline of the person (better viewed when zoomed in), while this gets filled with a clothes texture and color in Fig.~\ref{fig: ablation_target_mask}(j).
\subsection{Implementation Details}
\subsubsection{Pose Guided Warping Module (PGWM)}
The dense neural network in fashion landmark predictor network $\mathcal{F}$ contains 6 consecutive dense layers with 900, 800, 600, 500, 250, 100 nodes respectively, each having activation function tanh. Finally, the output layer has 12 nodes (for 6 fashion landmarks each with 2 coordinate values $x$ and $y$) and sigmoid as the activation function.

\subsubsection{Mask generator module (MGM)}
The architecture of MGM is that of an hourglass network~\cite{hourglass}. Generally, the processing of clothes requires identifying the different parts of them to establishing a semantic understanding of their structure. A well-known network that suits this requirement is the hourglass network~\cite{hourglass}. 

\begin{figure}[h]
	\centering
	\includegraphics[scale=0.13]{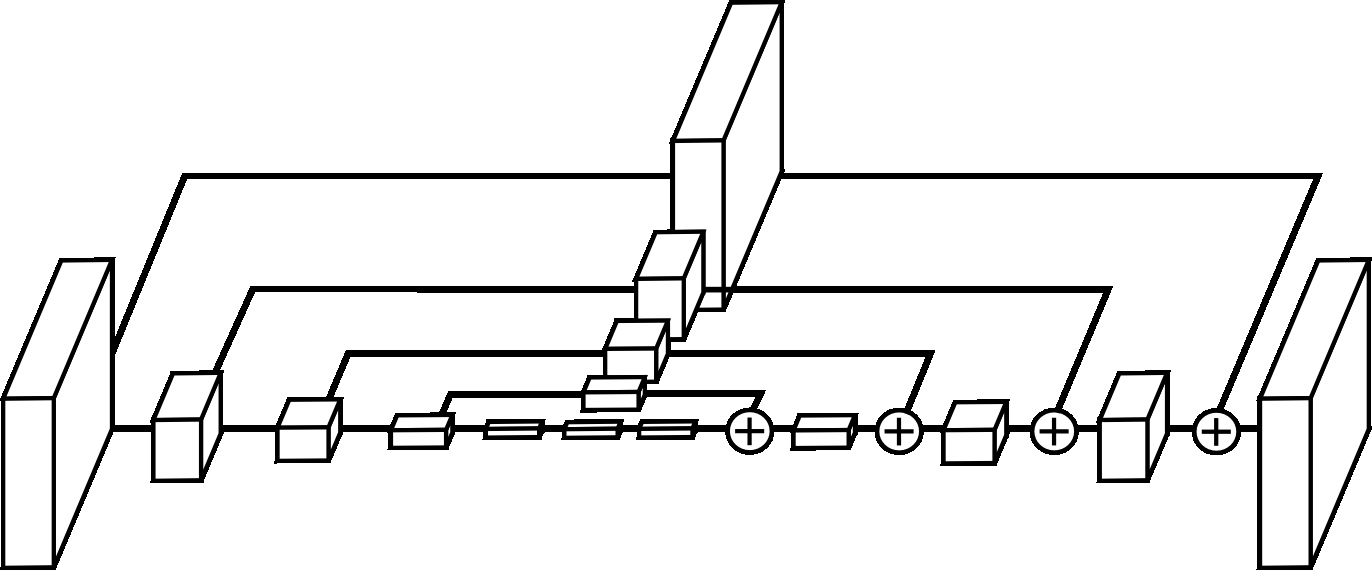}
	\caption{Overview of an hourglass network (taken from the original paper~\cite{hourglass}), where each block represents a residual module.}
	\label{fig: hourglass}
\end{figure}

An hourglass network is a CNN (Convolutional Neural Network) that captures features at various scales and is effective for analyzing spatial relationships among different parts of the input. Multiple of these hourglass networks can be stacked together with intermediate supervision for making it deeper. However, for our purpose, the stack size of 1 is found to be sufficient. An overview of the architecture of one hourglass network is given in Fig.~\ref{fig: hourglass}.
The network is called hourglass due to its top-down, bottom-up architecture, which matches with the shape of an hourglass. It uses convolution and max pooling layers to downscale features to a low resolution. After that, the top-down sequence of upsampling begins, and at each resolution, the corresponding features from downsampling parts are added as skip connections. However, it applies more convolutions on the features of skip connections and then does element-wise addition of the two sets of features.

\subsubsection{Image Synthesizer Module(ISM)}
\begin{figure}[!h]
	\centering
	\includegraphics[width=1\linewidth]{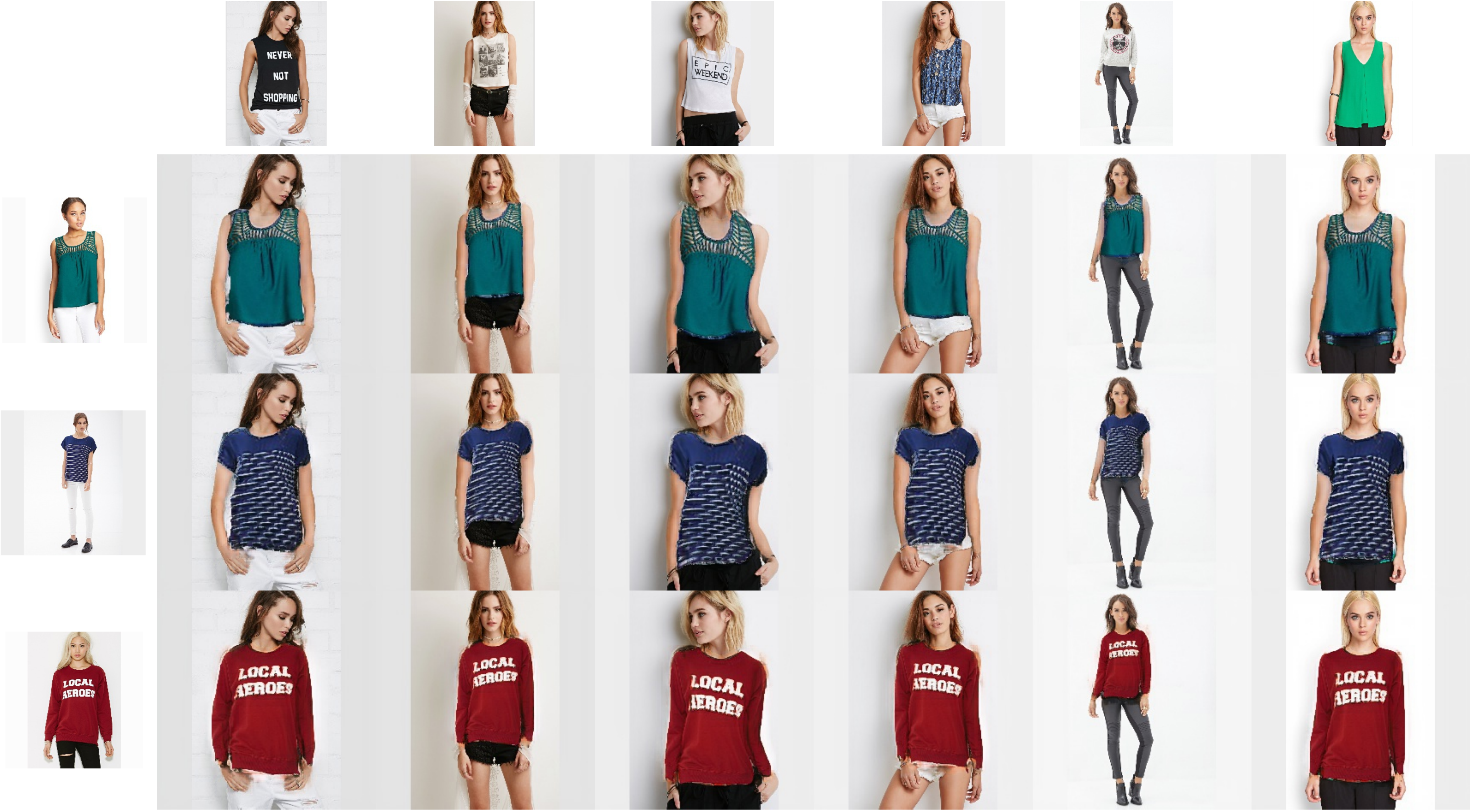}
	\caption{Our results on DeepFashion~\cite{deepfashion} dataset. Image at position (i,j) represents the result of VTON when person at $j^{th}$ column wears the clothes of the model at the $i^{th}$ row.}
	\label{fig: more_results_df1}
\end{figure}

\begin{figure}[h]
	\centering
	\includegraphics[scale=0.9]{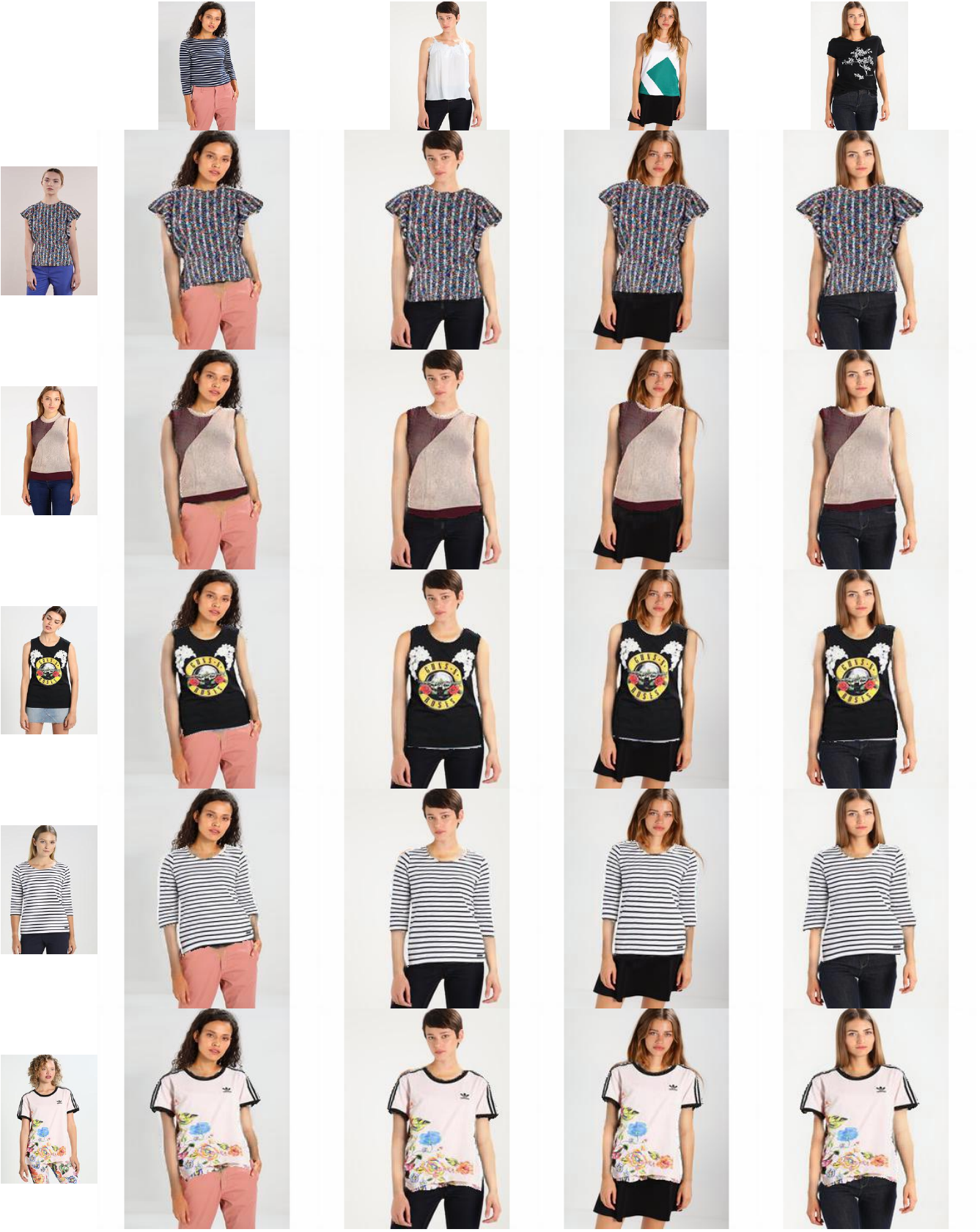}
	\caption{Our results on MPV~\cite{multiposevton} dataset. Image at position (i,j) represents the result of VTON when person at $j^{th}$ column wears the clothes of the model at the $i^{th}$ row.}
	\label{fig: more_results_mpv1}
\end{figure}

\begin{figure}[h]
	\centering
	\includegraphics[scale=0.36]{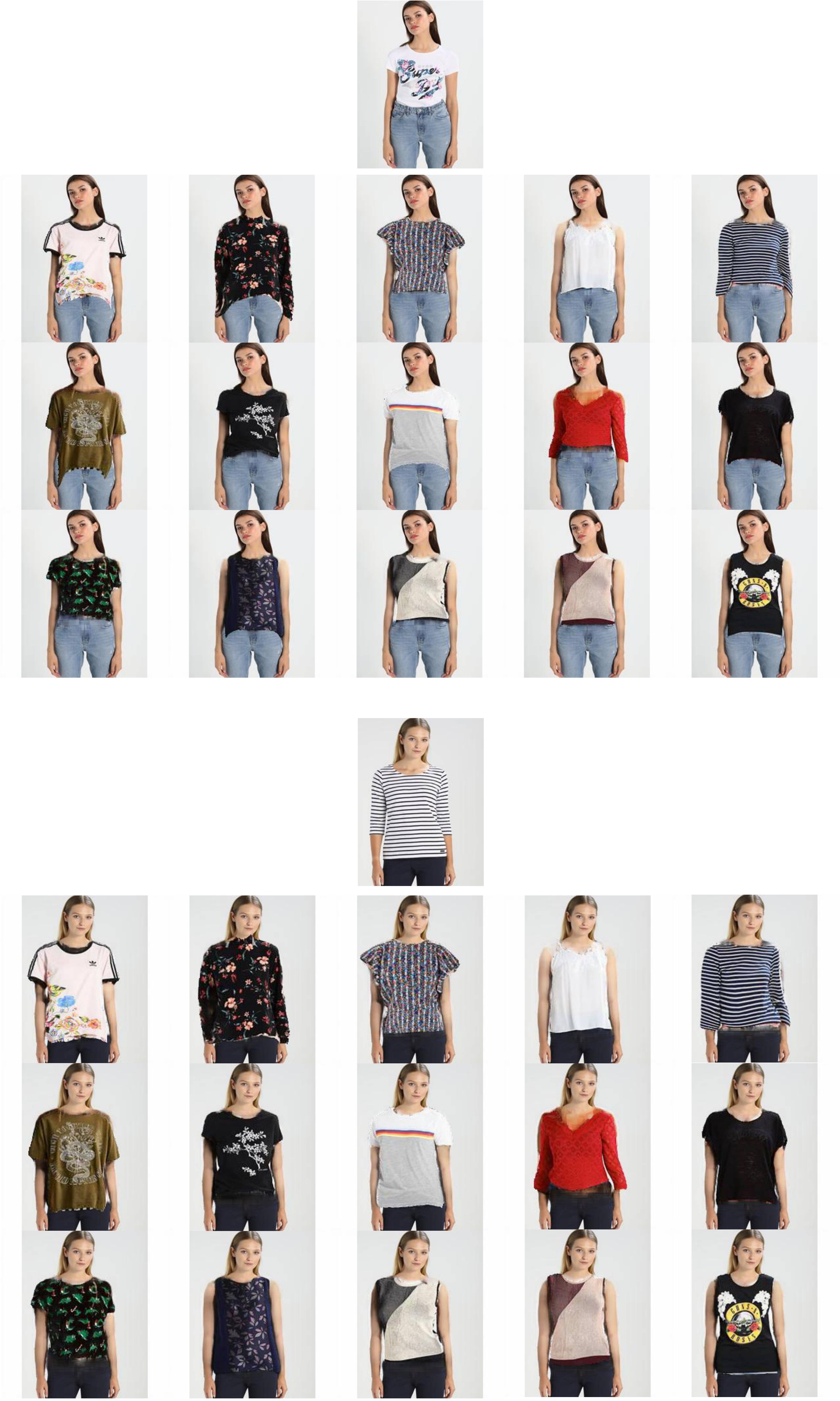}
	\caption{Our results on MPV dataset for different model and person combinations.}
	\label{fig: more_results_mpv2}
\end{figure}

\begin{figure}[h]
	\centering
	\includegraphics[scale=0.3]{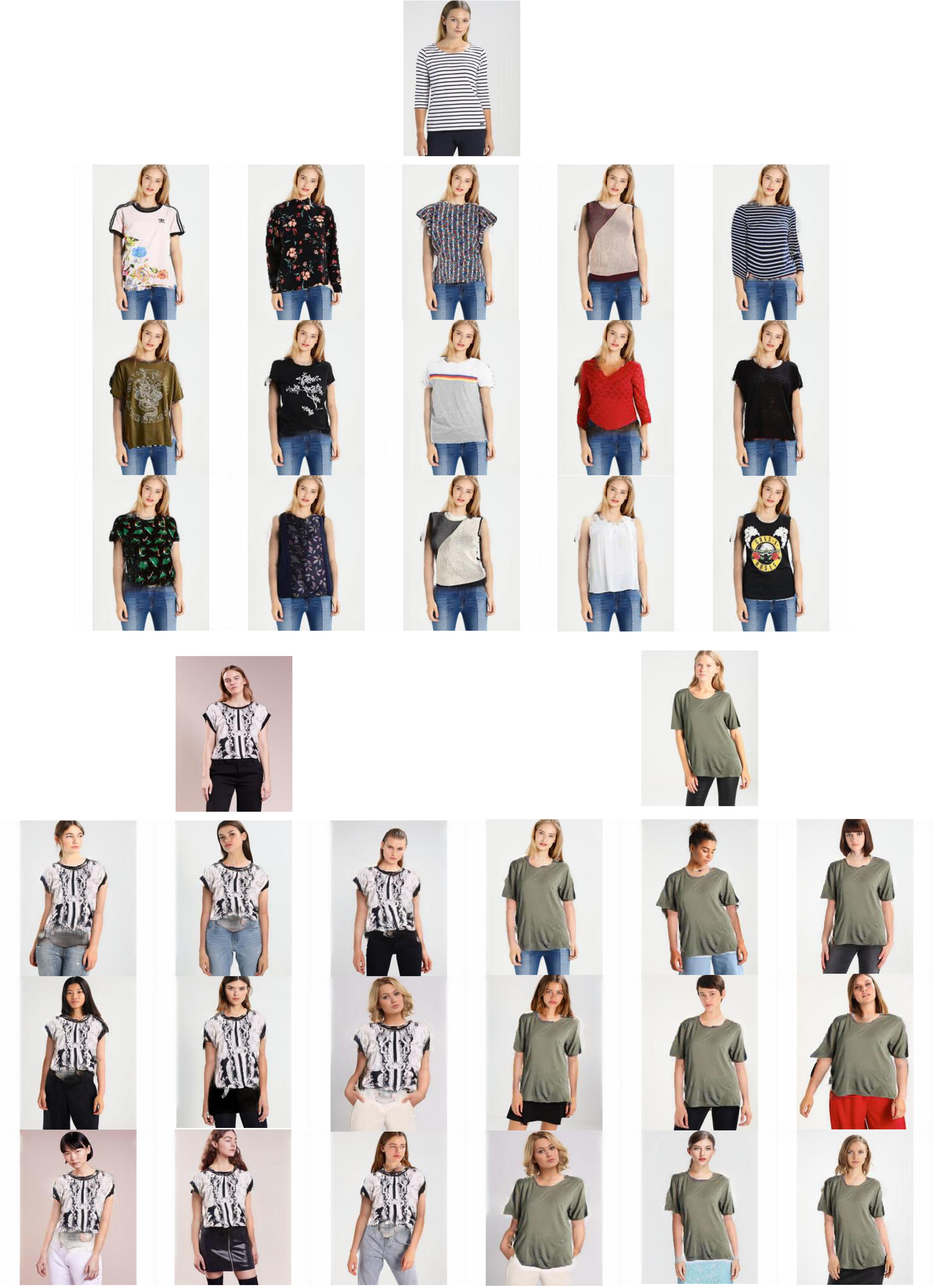}
	\caption{Our results on MPV dataset for different model and person combinations.}
	\label{fig: more_results_mpv3}
\end{figure}

The hourglass network in $G$ (the generator network of ISM) has a stack size of 1. The convolution layer generating $I_m$ in $G$ has a kernel of size 1 $\times$ 1 with $L_{1}$ regularization and sigmoid activation function. For training ISM, we alternate between 3 steps of generator training and 1 step of discriminator training. It is trained with the same settings of Adam as our PGWM. For DSSIM loss, the kernel size taken is 3 $\times$ 3. The discriminator $D$ is a patchGAN discriminator~\cite{pix2pix}. Instead of classifying the whole image as real or fake, it classifies each patch of the image, where the patch size is much smaller than the input image size. Hence pixels separated by more than a patch diameter gets modeled independently. This makes it work like a texture/style loss as discussed in~\cite{pix2pix}, which helps to keep better texture in the final output image. Existing works have shown the efficacy of patchGAN \cite{pix2pix},\cite{textureGAN} in image-based problems. For human parsing, we used the human parsing network proposed by~\cite{lipssl} pretrained on the LIP dataset~\cite{lipssl}. The dataset contains 19 part labels, 6 labels for body parts, and 13 for clothing categories. During our quantitative analysis we used the models of CP-VTON, VTNFP, MG-VTON pretrained on MPV~\cite{multiposevton} dataset. For VITON we used the VITON dataset pretrained model weights provided in its official implementations. 
Some more results of this work on DeepFashion and MPV datasets are shown in Figs.~\ref{fig: more_results_df1},~\ref{fig: more_results_mpv1},~\ref{fig: more_results_mpv2}, and~\ref{fig: more_results_mpv3}.
\clearpage
\bibliographystyle{spbasic}
\bibliography{bibliography}
\end{document}